\documentclass[conference]{IEEEtran}
\IEEEoverridecommandlockouts
\usepackage{cite}
\usepackage{amsmath,amssymb,amsfonts}
\usepackage{algorithmic}
\usepackage{graphicx}
\usepackage{hyperref}
\usepackage{caption}
\usepackage{textcomp}
\usepackage{xcolor}
\usepackage{amssymb, bm, mathtools}
\usepackage{siunitx}
\usepackage{subfig}
\usepackage{float}
\DeclareUnicodeCharacter{2212}{-}
\def\BibTeX{{\rm B\kern-.05em{\sc i\kern-.025em b}\kern-.08em
    T\kern-.1667em\lower.7ex\hbox{E}\kern-.125emX}}
    
\begin{document}

\title{Towards Understanding Underwater Weather Events in Rivers Using Autonomous Surface Vehicles 
\thanks{We greatly acknowledge the support of the NSF IUCRC 1939132 and the University of Pennsylvania's University Research Foundation Award.}
}

\author{\IEEEauthorblockN{Alice K. Li, Yue Mao, Sandeep Manjanna,  Sixuan Liu, \\ Jasleen Dhanoa, Bharg Mehta, \\ Victoria M. Edwards, Fernando Cladera Ojeda, M. Ani Hsieh}
\IEEEauthorblockA{\textit{GRASP Lab} \\
\textit{University of Pennsylvania}\\
Philadephia, USA}
\and
\IEEEauthorblockN{Maël Le Men, Eric Sigg, \\ Douglas J. Jerolmack, Hugo N. Ulloa}
\IEEEauthorblockA{\textit{Department of Earth and Environmental Science} \\
\textit{University of Pennsylvania}\\
Philadelphia, USA}
}



\maketitle

\begin{abstract}
Climate change has increased the frequency and severity of extreme weather events such as hurricanes and winter storms. The complex interplay of floods with tides, runoff, and sediment creates additional hazards --- including erosion and the undermining of urban infrastructure --- consequently impacting the health of our rivers and ecosystems. Observations of these underwater phenomena are rare, because satellites and sensors mounted on aerial vehicles cannot penetrate the murky waters. Autonomous Surface Vehicles (ASVs) provides a means to track and map these complex and dynamic underwater phenomena. This work highlights preliminary results of high-resolution data gathering with ASVs, equipped with a suite of sensors capable of measuring physical and chemical parameters of the river.  Measurements were acquired along the lower Schuylkill River in the Philadelphia area at high-tide and low-tide conditions. The data will be leveraged to improve our understanding of changes in bathymetry due to floods; the dynamics of mixing and stagnation zones and their impact on water quality; and the dynamics of suspension and resuspension of fine sediment. The data will also provide insight into the development of adaptive sampling strategies for ASVs that can maximize the information gain for future field experiments.
\end{abstract}


\section{Introduction}
Many of the largest cities have thrived because of their proximity to coastlines or rivers. The proximity to these large bodies of water provide a plethora of advantages such as drinking water, hydropower, shipping ports, transportation, waste disposal, and recreation \cite{balian2007freshwater, doyle2018source, macklin2015rivers, connor2015united, vorosmarty2010global}. Urban development, however, has negatively impacted these bodies of water, {\it e.g.}, increased pollution, interference with the delivery of water, sediment, and nutrients, among others \cite{best2019anthropogenic,opperman2017power,latrubesse2017damming,syvitski2005impact,walter2008natural,palmer2005standards, wohl2015science}. In addition, increasingly intense rainfall events driven by climate change have altered their hydroclimates, creating larger floods and longer dry spells. Flooding can alter the shape and size of river channels \cite{slater2013imprint, slater2015hydrologic, slater2019river}. For tidally-influenced rivers like the lower Schuylkill River in Philadelphia, PA, rising sea levels further enhance flood risk and ecosystem disruption \cite{yang2015estuarine}. Such flooding can cause erosion that undermines infrastructures such as bridges and piers \cite{garcia2008sedimentation}, as well as the re-suspension of PCB-contaminated legacy sediments \cite{du2008source}. 

The movement of water, solutes and sediments through estuaries is governed by a complex interplay among river currents, tides, buoyancy, and bathymetry \cite{geyer2014estuarine, egan2020waves}. Mud deposition resulting from changes in flow or water chemistry \cite{winterwerp2002flocculation} can shrink the channel cross section, enhancing flood risk \cite{garcia2008sedimentation, slater2021nonstationary} and suffocating bottom-dwelling organisms \cite{soulsby2001fine}.  These ``underwater weather'' events, as described in \cite{monroe2002underwater}, a term we use to summarize flow and sediment dynamics occurring under the water surface, are therefore critically important for understanding how to protect river health. Yet, observations of the relevant meteorological variables, if available, are typically sparse in terms of both spatial and temporal coverage. This is because sensors are mounted either in stationary locations, or on large boats where surveys are infrequent.  Based on the need to improve our understanding of a river's response to extreme weather events, the goal of this work is to document underwater weather events in the lower Schuylkill River in Philadelphia for the first time. Data gathered will be used to determine the cumulative effects these weather events have on infrastructure and water quality. In this work, we leverage the mobility of ASVs to conduct high resolution surveys of the river structure and its dynamics. This data will also be used to develop long term adaptive sampling strategies for ASVs, data-driven modeling of the system dynamics, and gaining new physical insights.

\section{Related Work}
Autonomous surface vehicles are increasingly being used to acquire a variety of in-situ water measurements from oceans, rivers, and lakes. Hitz et al. provide the design of a customized ASV used to monitor toxic algae blooms \cite{hitz2012autonomous}. Their custom-made ASV and winch system weights \SI{120}{kg} and allows for lake measurements at depths of \SI{130}{m}. The ASV detailed in \cite{Fornai} acquires water samples up to \SI{50}{m} in depth, allowing for measurements of physical water parameters along the water column. In \cite{Manjanna2018}, chlorophyll density is mapped, as an exploring ASV determines regions of interest for a data collecting ASV to acquire water samples. Partial pressures and pH sensors were used on board Wave Glider autonomous surface vehicles to sample in the central California upwelling system \cite{chavez2018measurements}. Dunbabin et al. uses a \SI{16}{ft} long solar powered ASV for water quality and greenhouse gas monitoring \cite{dunbabin2009autonomous}. Other works detailing ASVs with sensor suites for water quality and bathymetry measurements can be found in \cite{valada2014development, Desa, Madeo2020ALU, Simmerman, demetillo2019real, sukhatme2007design} and in a review \cite{Manley2008UnmannedSV}. 
In \cite{Velasco}, Velasco et al. designed an ASV with a similar sensor suite to that described in the current work. However, they focus on highlighting their robust mechanical design for operating around strong waves of the north coastline of Peru. Most works also focusing on the mechanical design of their ASVs opt for a catamaran design, similar to that of the ASVs in the current work, allowing for operation in shallow waters and increased stability against waves \cite{Ferreria,Wang}. Yang et al. use unmanned surface vehicles to understand the impacts that extreme weather has on seabeds surrounding Taiwan; however, their vehicles are remote controlled rather than autonomous  \cite{yang2011multifunctional}. Li et al. use surface vehicles to acquire data to better understand and predict floods in the case of extreme weather events in water channels. In their work, they only measure vertical profiles of horizontal flow velocity with an ADCP sensor \cite{li2018weather}.

As evidenced by the literature, there is growing interest in using ASVs to acquire water quality and at depth measurements.  Nevertheless, to the authors' knowledge, none directly addresses leveraging autonomy for improving our understanding of river and sediment dynamics resulting from extreme weather events, nor establish correlations between measured parameters.  In this work, we describe our on-going efforts towards the development of long-term autonomous monitoring of underwater weather events in riverine environments. 

\vspace{-4mm}

\begin{figure}[H]
    \centering
    \includegraphics[width=\linewidth]{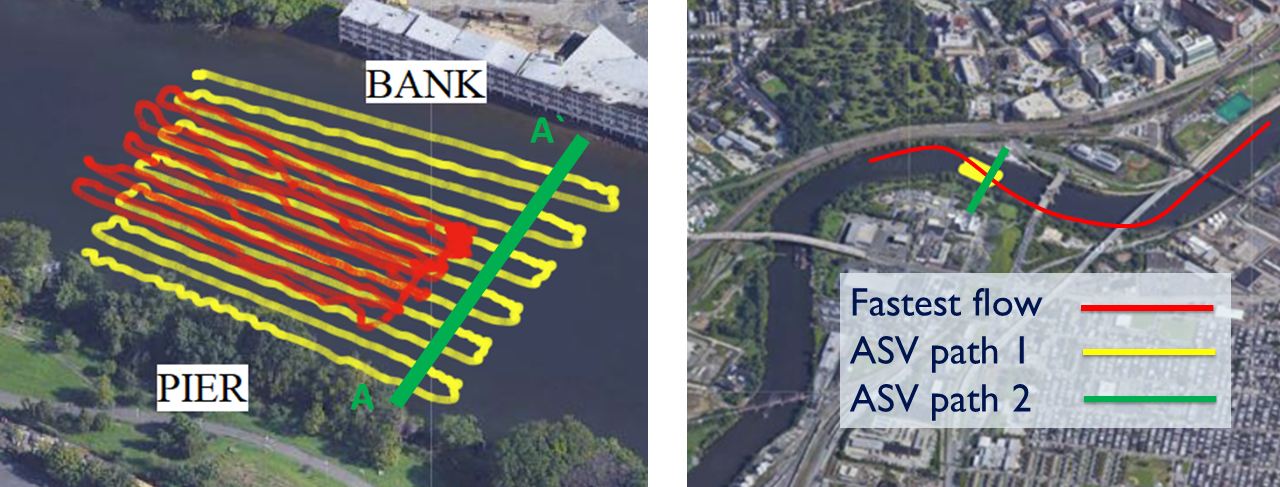}
    \caption{\textit{Left} Example paths taken by the ASVs, along and across the Schuylkill River. Red represents an example lawn-mower  coverage pattern by the custom made ASV. Measurements acquired by this robot include pH, temperature, nitrate, pressure, barometric pressure, oxidation reduction potential (ORP), and chlorophyll-a fluorescence, sediment concentration, and bathymetry. Yellow represents coverage of a larger region by the Clearpath Heron, acquiring bathymetry measurements. Green represents the approximate path taken by the custom made ASV across the river, for 2D riverbed reconstruction. \textit{Right} Lawn-mower pattern depicted by a yellow rectangle, and green path across the river. Measurement locations are situated between two meanders in the river, possibly giving rise to asymmetric riverbed shape. Figure background courtesy of Google Maps.}
    \label{paths_over_river}
\end{figure} 

\begin{figure}[H]
    \centering
    \subfloat[Clearpath Heron ASV with RTK base station.]{\includegraphics[width = 0.7 \linewidth]{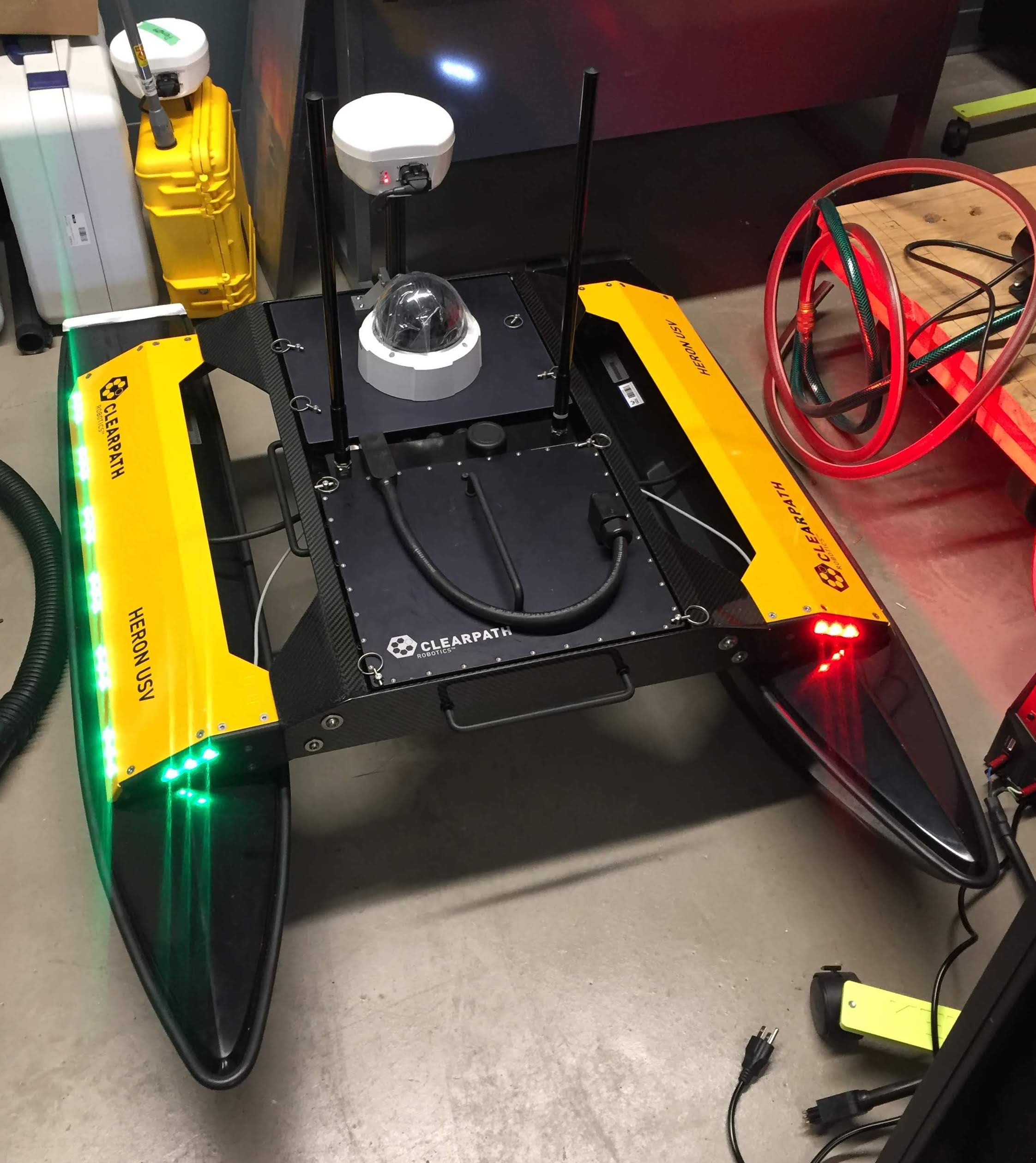}\label{fig_Heron}}
    \hspace{0.5cm}
    \subfloat[Customized lightweight ASV, which holds LISST-ABS, In-Situ Aqua Troll 600 sensor, and  BlueRobotics bathymetry sensor.]{\includegraphics[width=0.7\linewidth]{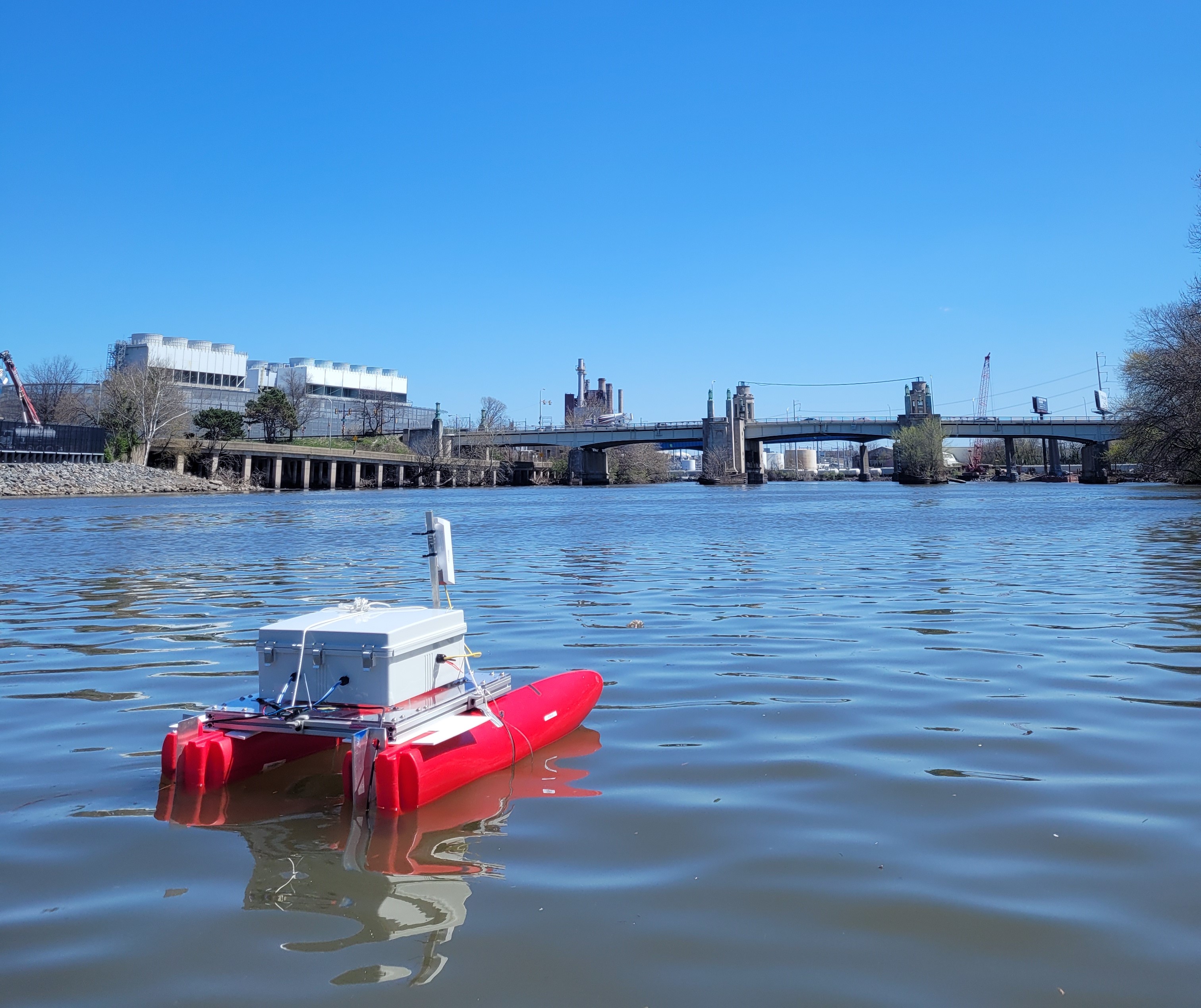}\label{lightweight_asv}}
    \hspace{0.5cm}
    \subfloat[Sensor suite - LISST-ABS sensor (top) In-Situ Aqua Troll 600 sensor. (bottom)]{\includegraphics[width=0.7\linewidth]{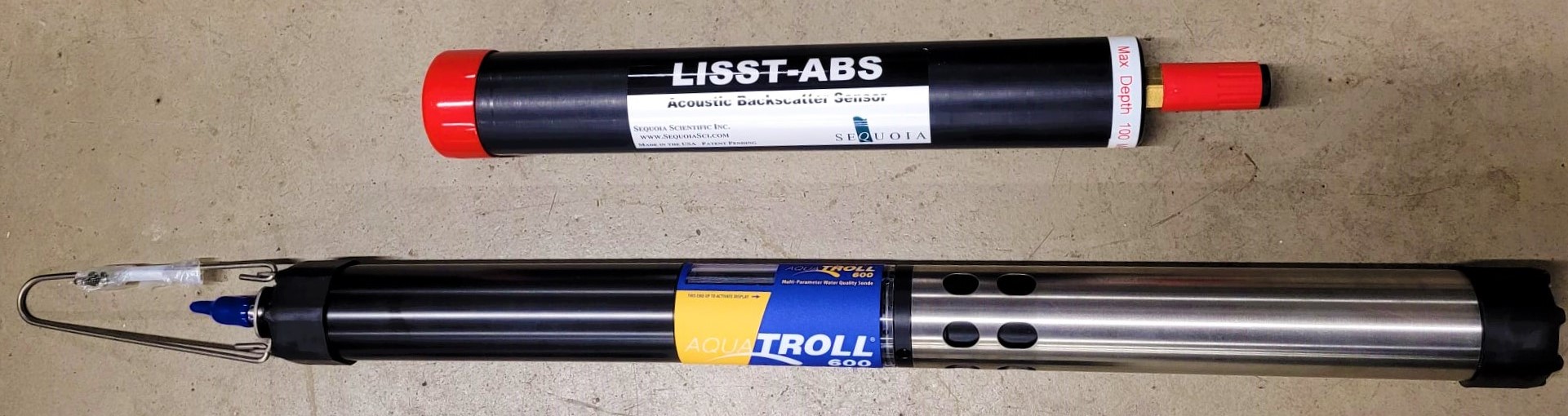}\label{sensor_suite}}
    \hspace{0.5cm}
    \subfloat[Tritech Micron Bathymetry sensor on board the Clearpath Heron ASV.]{\includegraphics[width = 0.7 \linewidth]{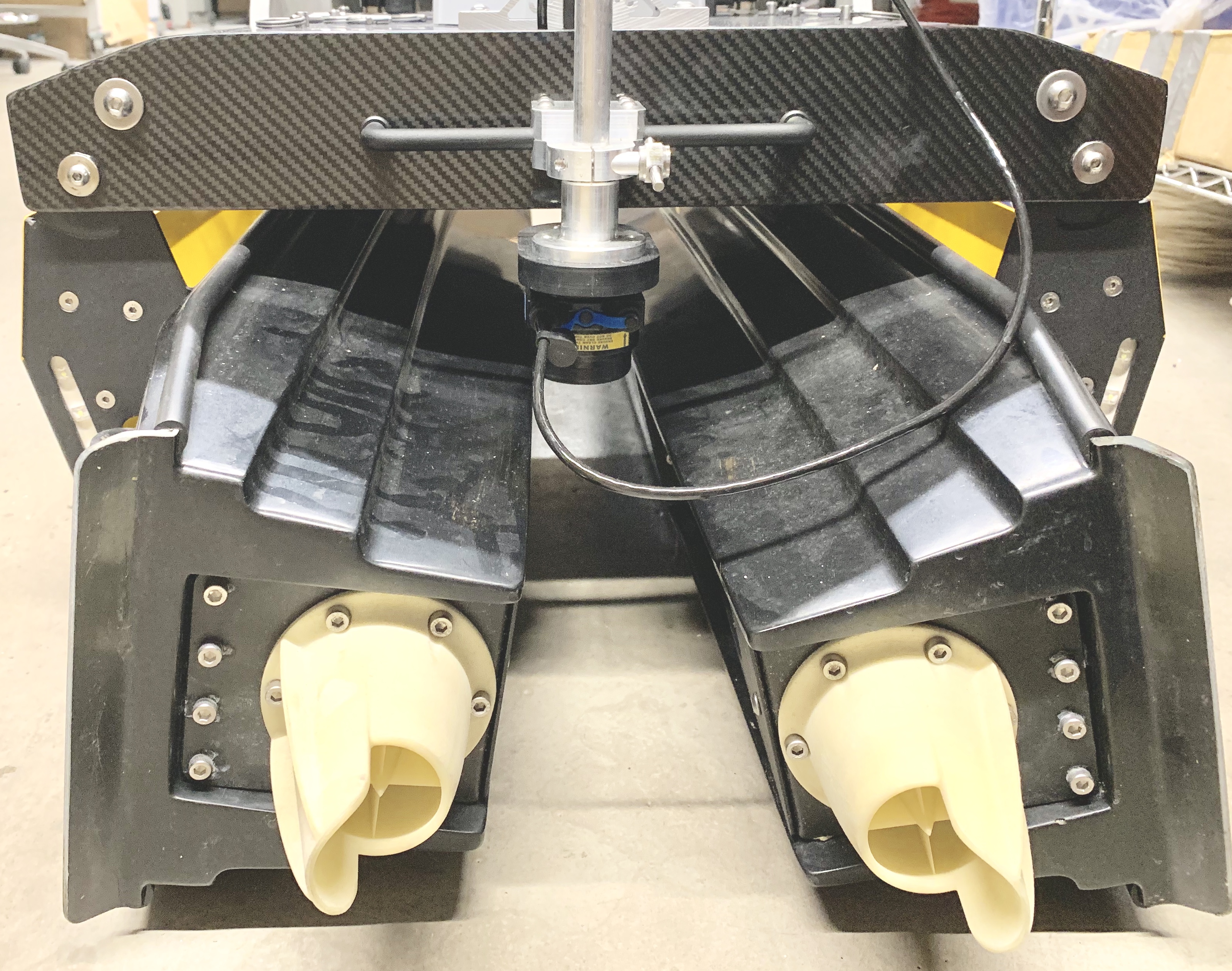}\label{fig_echosounder}}
    \caption{Experimental setup.}
\end{figure}
\section{Experimental Methods}
In this work, we employed two types of ASVs -- a Clearpath Heron Autonomous Surface Vehicle with a bathymetry sensor and a custom made ASV with a suite of sensors -- to carry out surveys in the Schuylkill river in Philadelphia. Figure \ref{paths_over_river} shows the paths traveled by the Clearpath Heron (yellow) and custom made ASV (red). Figure \ref{paths_over_river} shows the region of coverage from a zoomed out view. The custom made ASV, which had the sensor suite on board, to be described in further detail below, surveyed a segment of the river under high-tide and low-tide conditions, covering a spatial region of around \SI{90}{m} $\times$ \SI{40}{m} enclosed by GPS coordinates $(39.94364, -75.19973)$, $(39.94403, -75.19943)$, $(39.94356, -75.19856)$, and $(39.9432, -75.19883)$. The Clearpath Heron, with greater battery capacity covered a larger spatial region of around \SI{100}{m} $\times$ \SI{75}{m}, and was used to acquire bathymetry data. The custom made ASV was also used to measure bathymetry measurements perpendicular to the flow of the river, to generate 2D cross-sectional plots.

The Clearpath Heron is a differentially driven ASV with a catamaran design. It has a built-in GPS, a 6 DOF inertial measurement unit (IMU) and a camera (Figure \ref{fig_Heron}). To improve GPS measurements, we have installed a NovAtel Real Time Kinematic (RTK) GPS on board the ASV. Coupling this GPS with an RTK base station provides localization of $\pm$ \SI{1}{cm} accuracy. The custom made ASV, Figure \ref{lightweight_asv}, is also differentially driven, with a catamaran design consisting of two pontoons and two T200 BlueRobotics thrusters. It has an Intel NUC i7, a Pixhawk 4 controller, and on-board Pixhawk GPS with accuracy $\pm$ \SI{2}{m}. A TP-link router along with the NUC creates an ad-hoc network. This allows us to communicate with the NUC to control the ASV via ROS over WiFi, while it is in the river. The ASV can also be remotely operated via a \SI{2.4}{GHz} RC controller at a range of around \SI{500}{m}. 

The suite of sensors that are mounted on the custom ASVs include (see Figure \ref{sensor_suite}): (1) a BlueRobotics Ping Sonar Altimeter and an Echosounder bathymetry sensor with accuracy of \SI{0.5}{\%} of the range, with sensing range of \SI{0.5}{m} - \SI{70}{m}. This bathymetry sensor samples at \SI{10}{Hz}. (2) In-Situ Aqua Troll 600 sensor, capable of measuring numerous physical and chemical parameters including:  pH, temperature, nitrate, pressure, barometric pressure, oxidation reduction potential (ORP), and chlorophyll-a fluorescence, sampling at \SI{0.5}{Hz}. (3) LISST-ABS sediment concentration sensor, a light-scattering acoustic sensor for measuring particle concentration, sampling at \SI{1}{Hz}. On board the Clearpath Heron is a Tritech Micron Echosounder bathymetry sensor, capable of measuring up to a \SI{50}{m} water depth with a resolution of \SI{1}{mm} (Figure \ref{fig_echosounder}) is mounted at the rear of the Clearpath Heron ASV. The use of non-identical bathymetry sensors is due to availability.

For this work, we ran the ASVs autonomously along a boustrophedon pattern, {\it i.e.}, a lawn mower pattern, in the region of interest, as depicted in Figure \ref{paths_over_river}. Physical and chemical measurements with corresponding GPS coordinates were acquired to construct spatial maps of these phenomena. Measurements were acquired under two tidal conditions, as data could potentially be extrapolated to predict river conditions during floods if salient trends exist. Data taken within $1.5$ hours of peak high and low tide conditions are assumed to be influenced by the affects of high and low tidal conditions respectively. Low tide occurred on  $2022-08-09$ at $12:50$ PM at a height of \SI{0.5}{ft}, and high tide occurred on $2022-08-03$ at $11:46$ AM at a height of \SI{5.8}{ft}. To remove as much variability from ambient temperature conditions, all data was collected at similar times of the day -- 2 hours within noon. 3 data sets were collected for low tide and 2 data sets were collected for high tide. The number of data sets were limited by weather and time for data acquisition. Ambient temperatures for low and high tidal conditions were around \SI{33}{^\circ C} - \SI{34}{^\circ C} and \SI{31}{^\circ C} - \SI{32}{^\circ C} respectively. Point measurements were acquired to generate spatial maps of pH, temperature, nitrate, pressure, barometric pressure, ORP, and chlorophyll-a fluorescence, sediment concentration, and depth over the region. Only plots with significant relevance and conclusive results are included. 

\begin{figure}[htb]
    \centering
    \includegraphics[width=1\linewidth]{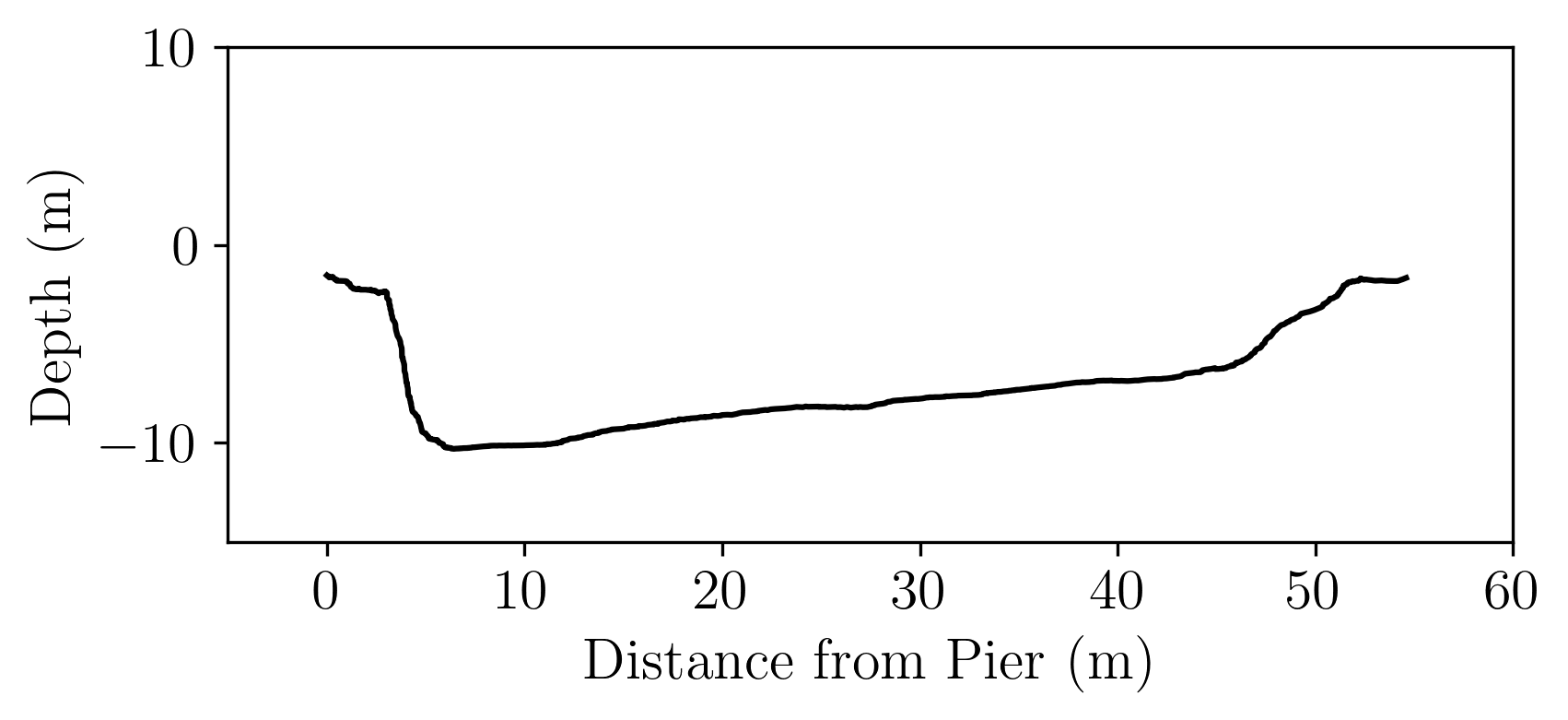}
    \caption{2D reconstruction of the riverbed.}
    \label{2d_reconstruction}
    \vspace{0mm}
    \centering
    \includegraphics[width=1\linewidth]{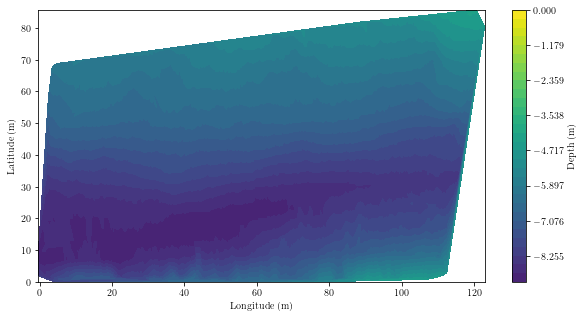}
    \caption{A triangulation based bi-linear interpolation approach was used to interpolate between point measurements acquired along the yellow path, shown in Figure 1, to produce a spatially continuous map of depth data.}
    \label{depth_heron}
\end{figure}

\section{Results and Discussion}
2D cross-sectional areas of the riverbed were constructed by guiding the custom ASV perpendicular to the flow of the river. Two sets of data were acquired and averaged, via a sliding window to generate Figure \ref{2d_reconstruction}. The left hand side of the plot is closest to the pier shown in Figure \ref{paths_over_river}, while the right hand side is closest to the bank. Figure \ref{depth_heron} shows an interpolated spatial plot representing depth across the river, using the Clearpath Heron, the yellow path in \ref{paths_over_river}. Given that low latitude represents data closer to the pier, it is in agreement with Figure \ref{2d_reconstruction}, which shows that the riverbed is deeper closer to the pier. It should be noted that just before and just after the pier, there is a bend in the river. Such meandering can give rise to the non-symmetric riverbed structure -- the greater velocities of the river at the outer edge of the bend causes greater erosion, and therefore gives rise to greater riverbed depth. While the riverbed depth should end at \SI{0}{m} on each side, we are limited in mobility close to the banks, as the sensor heads are positioned around \SI{15}{cm} below the water surface.

A triangulation based bi-linear interpolation approach was used to interpolate between point measurements acquired along the paths shown in Figure \ref{paths_over_river}, to produce a spatially continuous map. The red path is that of the custom-made ASV; the yellow is that of the Clearpath Heron. The final coverage regions are not identical because the ASVs motions are affected by the river currents. Since data rates for all sensors are different, data were synchronized to match corresponding timestamps. The spatial measurements were converted from GPS coordinates to meters, that covered a rectangular region of 90 m x 40 m. They were also rotated for easier readability and alignment with the final coordinate system.

\begin{figure}[H]
    \centering
\subfloat[High-tide at 11:47AM on 08-03-2022.]{\includegraphics[width = 0.8\linewidth]{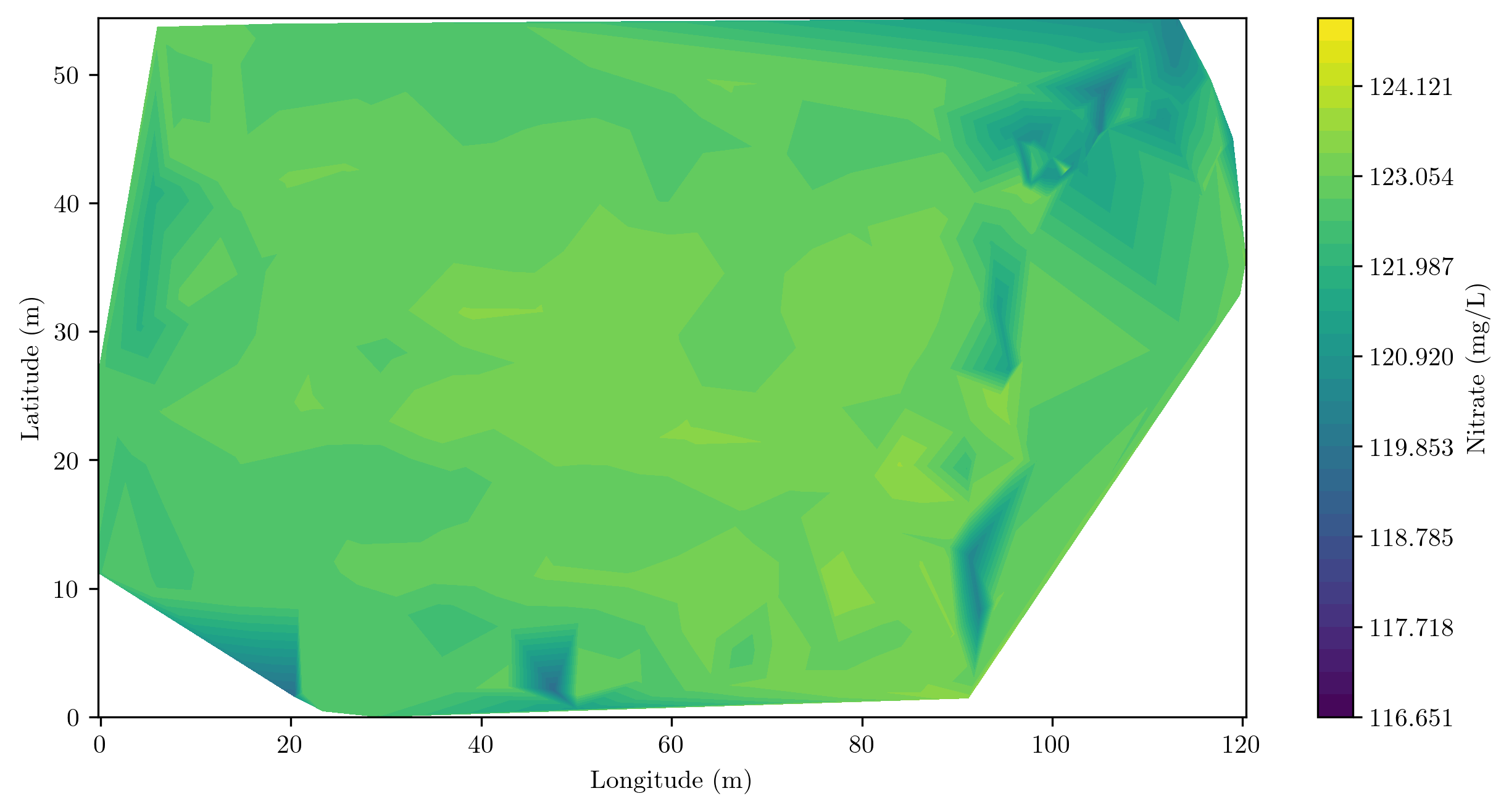}\label{nitrate1}}		\hspace{0.1cm}
\subfloat[High-tide at 1:34AM on 08-03-2022.]{\includegraphics[width = 0.8 \linewidth]{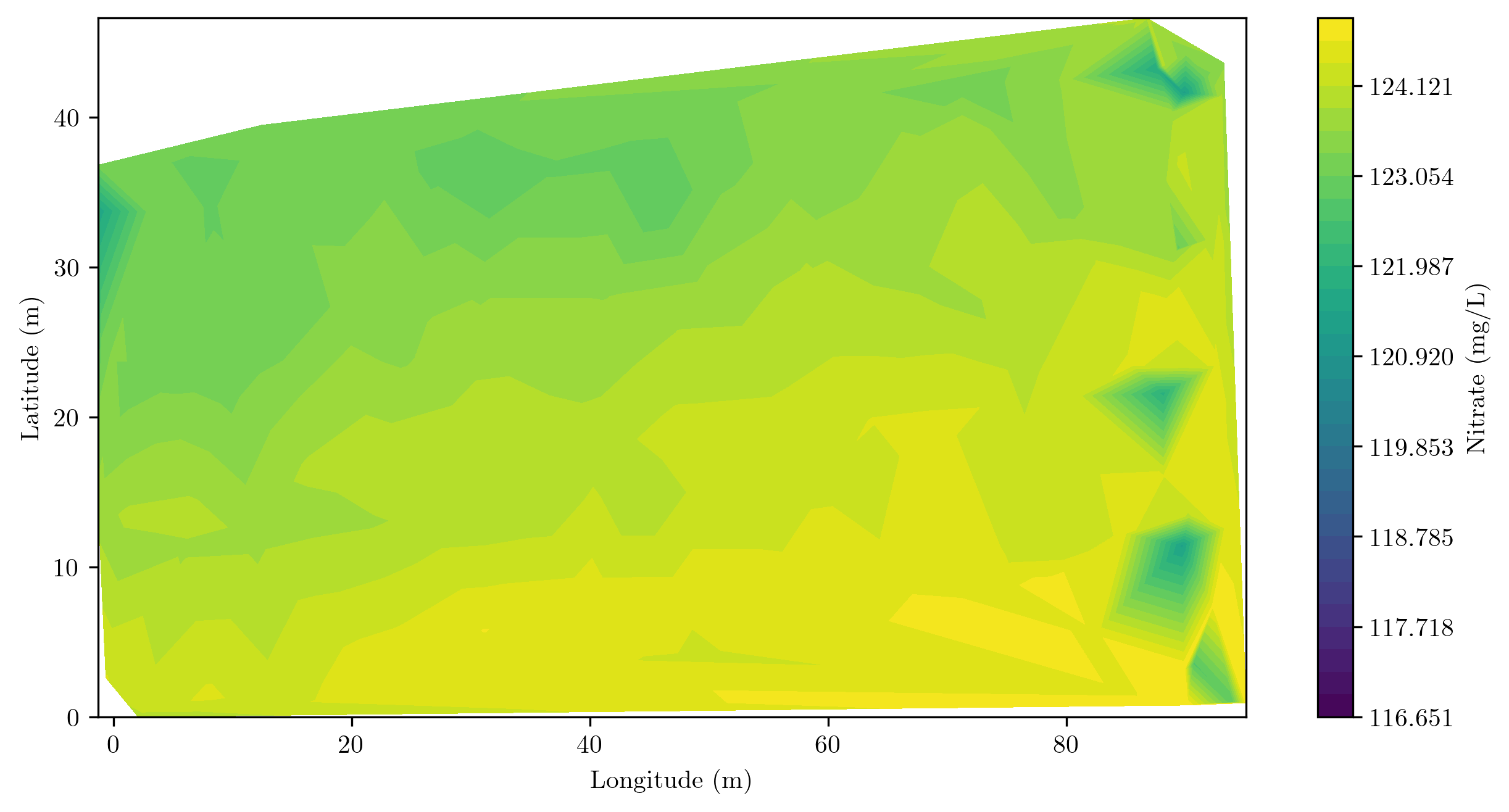}\label{nitrate2}}		\hspace{0.1cm}
\subfloat[Low-tide at 11:07AM on 08-09-2022.]{\includegraphics[width =0.8 \linewidth]{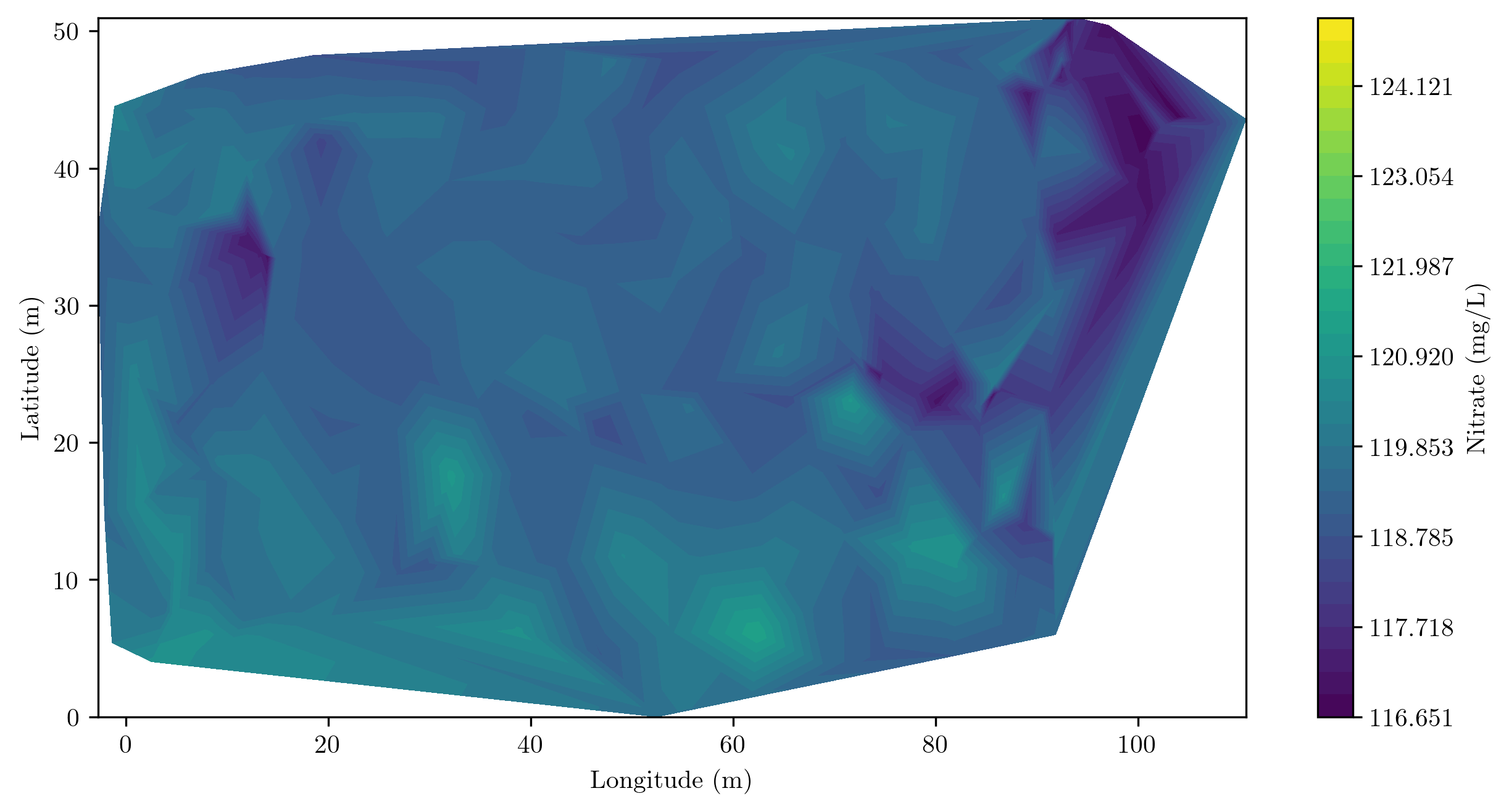}\label{nitrate3}}		\hspace{0.1cm}
\subfloat[Low-tide at 12:03PM on 08-09-2022.]{\includegraphics[width = 0.8 \linewidth]{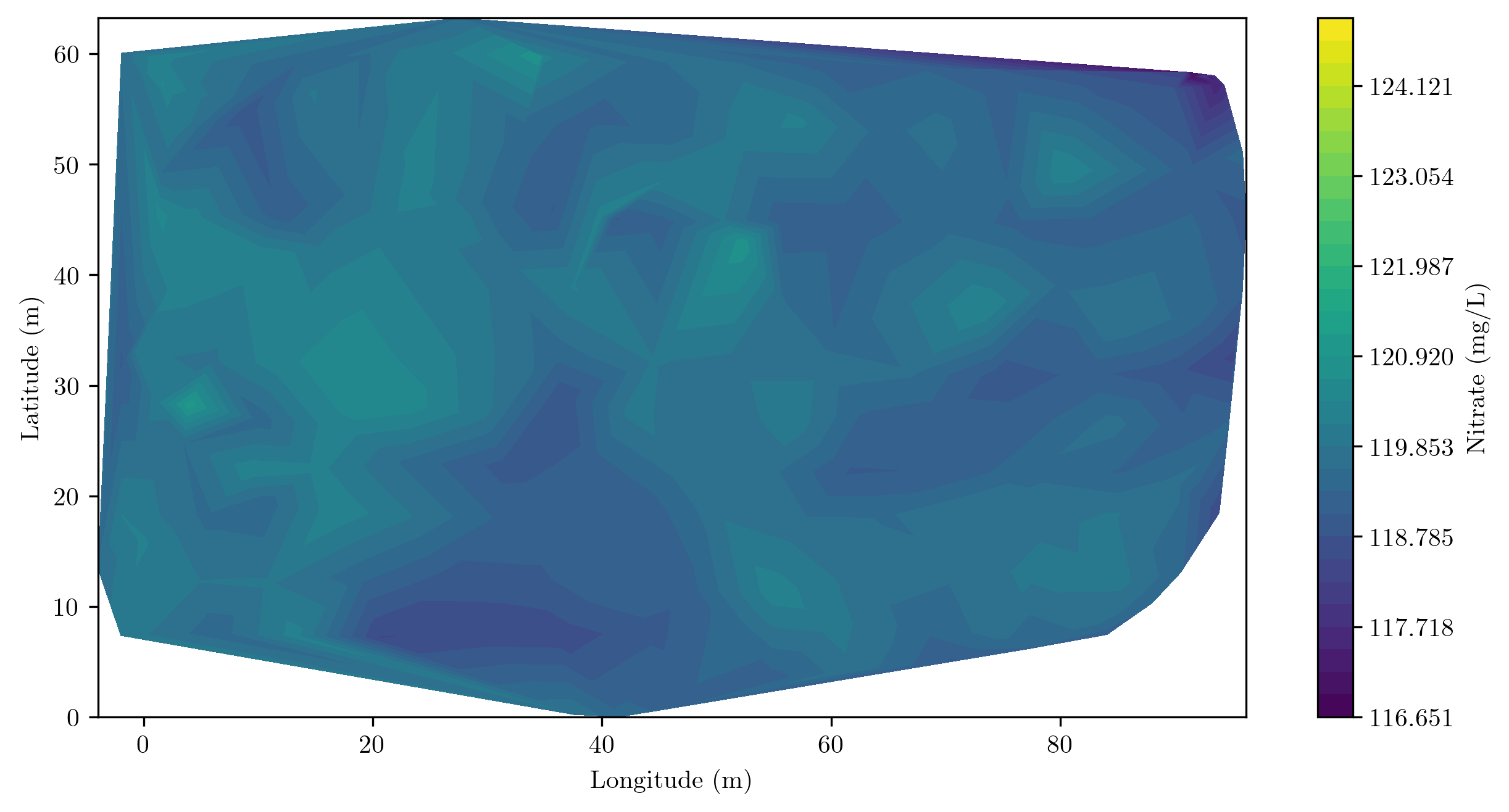}\label{nitrate4}}		\hspace{0.1cm}
\subfloat[Low-tide at 12:19PM on 08-09-2022.]{\includegraphics[width = 0.8 \linewidth]{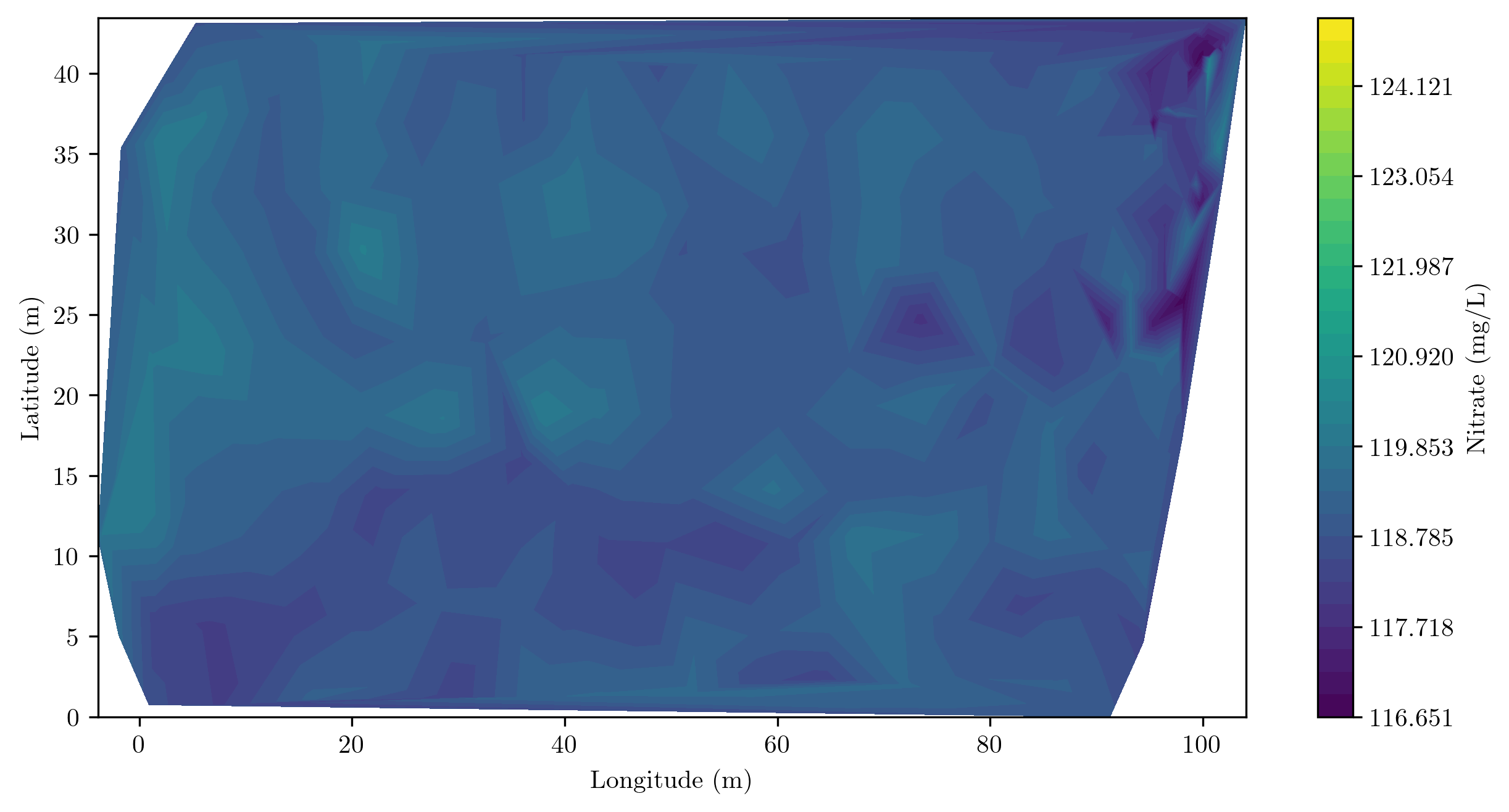}\label{nitrate5}}		\hspace{0.1cm}
\caption{Interpolated nitrate measurements.}
\label{nitrate_all}
\end{figure}

\begin{figure}[H]
    \centering
\subfloat[High-tide at 11:47AM on 08-03-2022.]{\includegraphics[width = 0.8\linewidth]{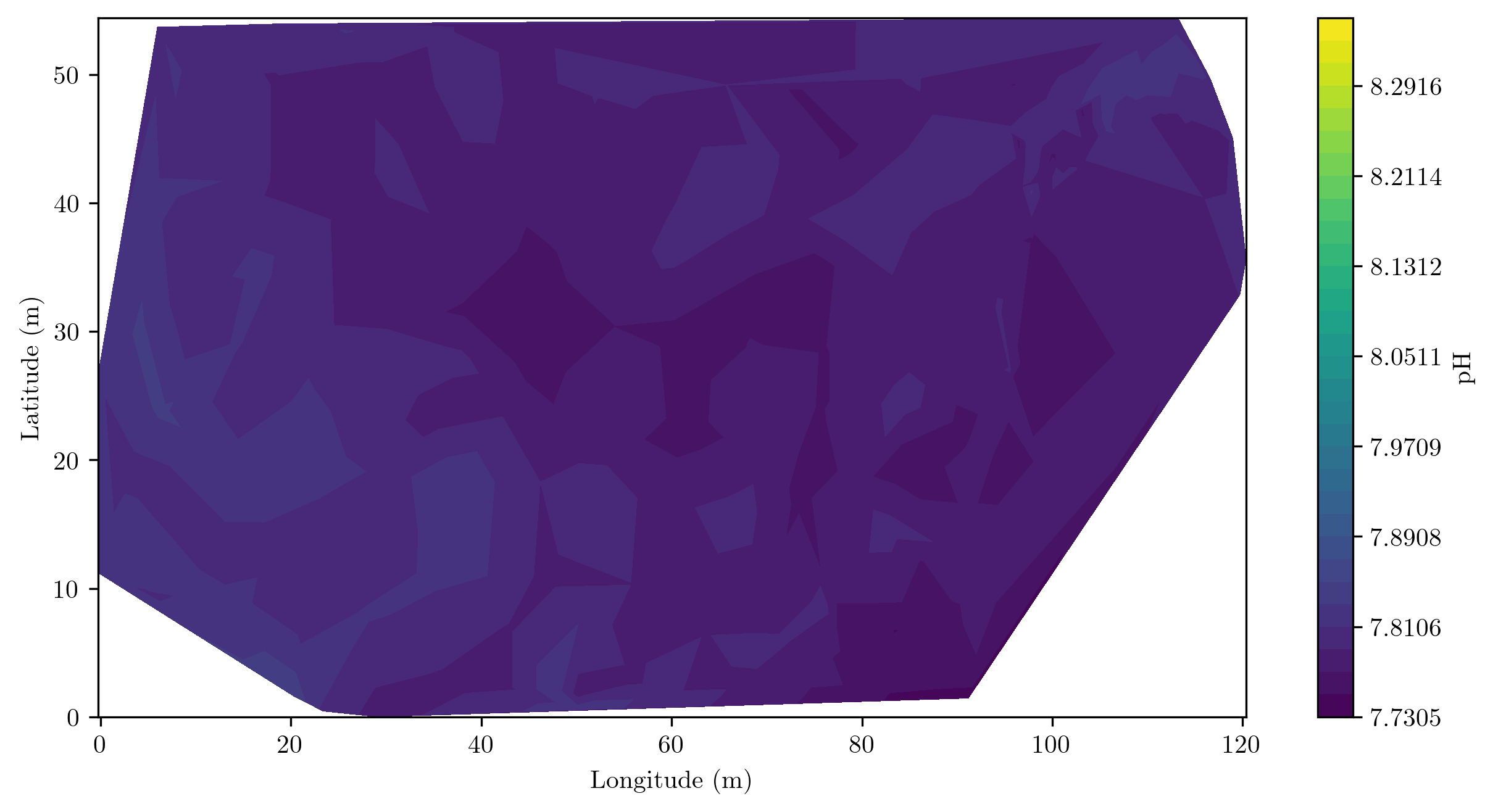}\label{pH1}}		\hspace{0.1cm}
\subfloat[High-tide at 1:34PM on 08-03-2022.]{\includegraphics[width = 0.8 \linewidth]{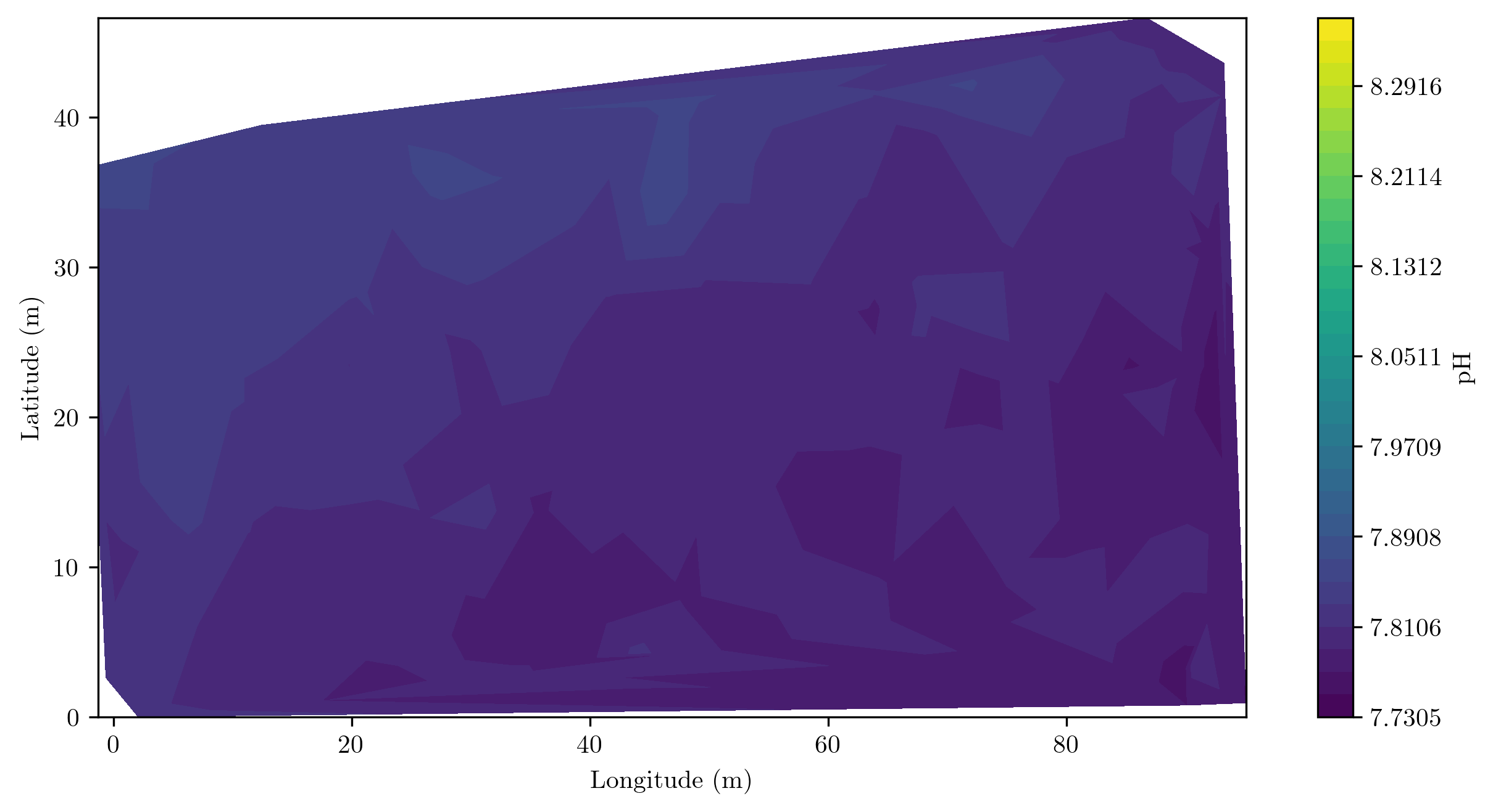}\label{pH2}}		\hspace{0.1cm}
\subfloat[Low-tide at 11:07PM on 08-09-2022.]{\includegraphics[width =0.8 \linewidth]{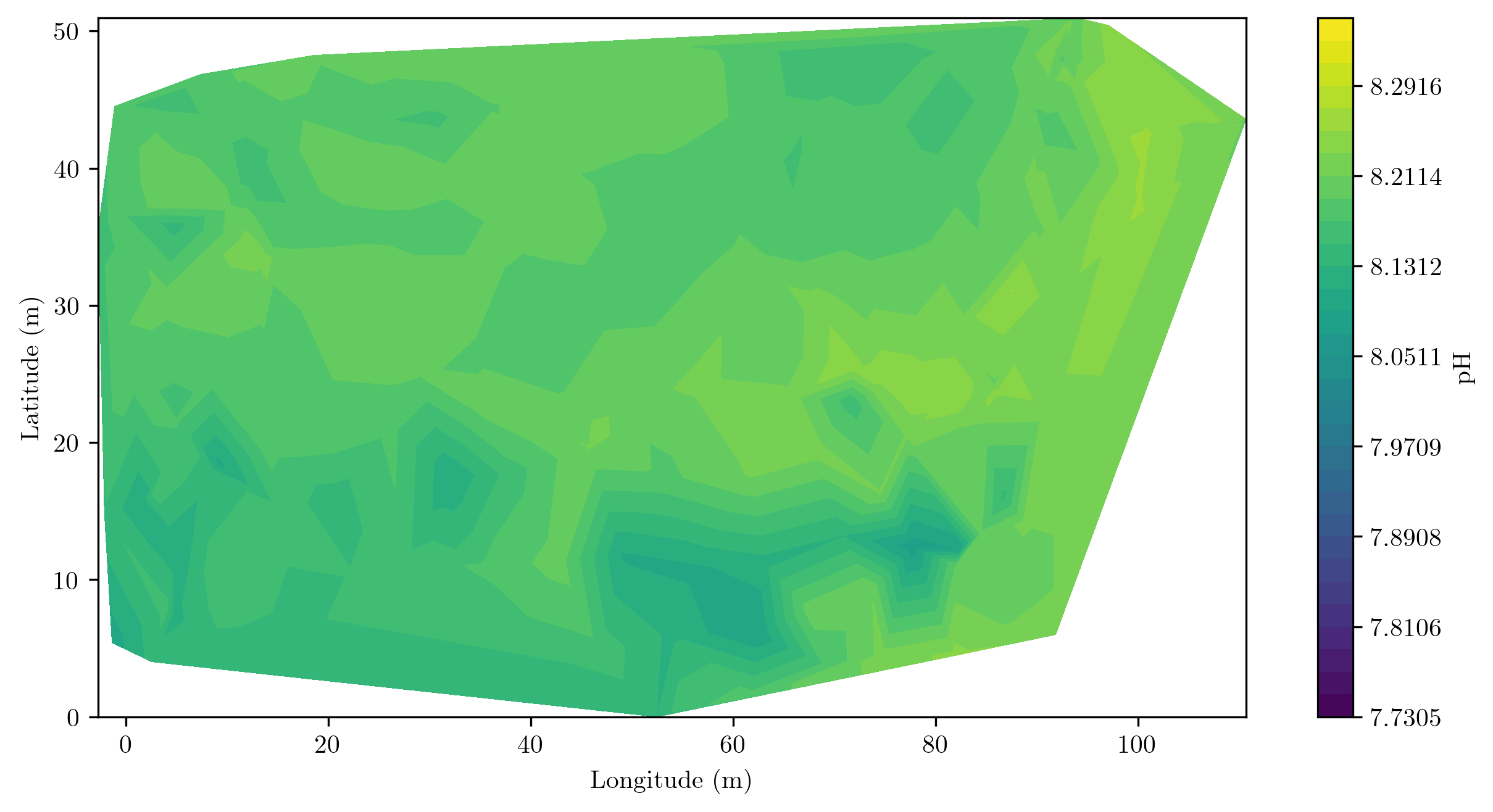}\label{pH3}}		\hspace{0.1cm}
\subfloat[Low-tide at 12:03PM on 08-09-2022.]{\includegraphics[width = 0.8 \linewidth]{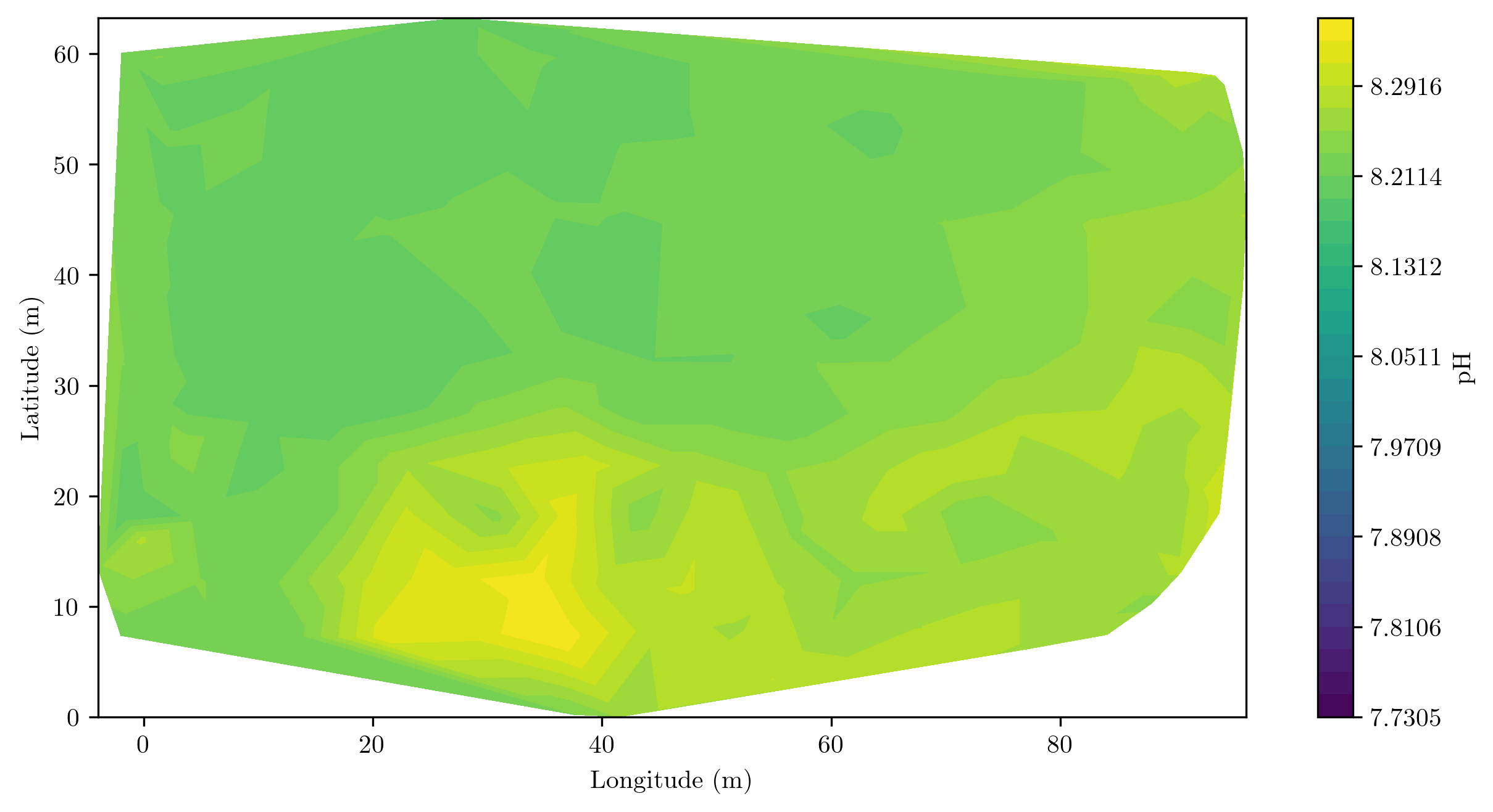}\label{pH4}}		\hspace{0.1cm}
\subfloat[Low-tide at 12:19PM on 08-09-2022.]{\includegraphics[width = 0.8 \linewidth]{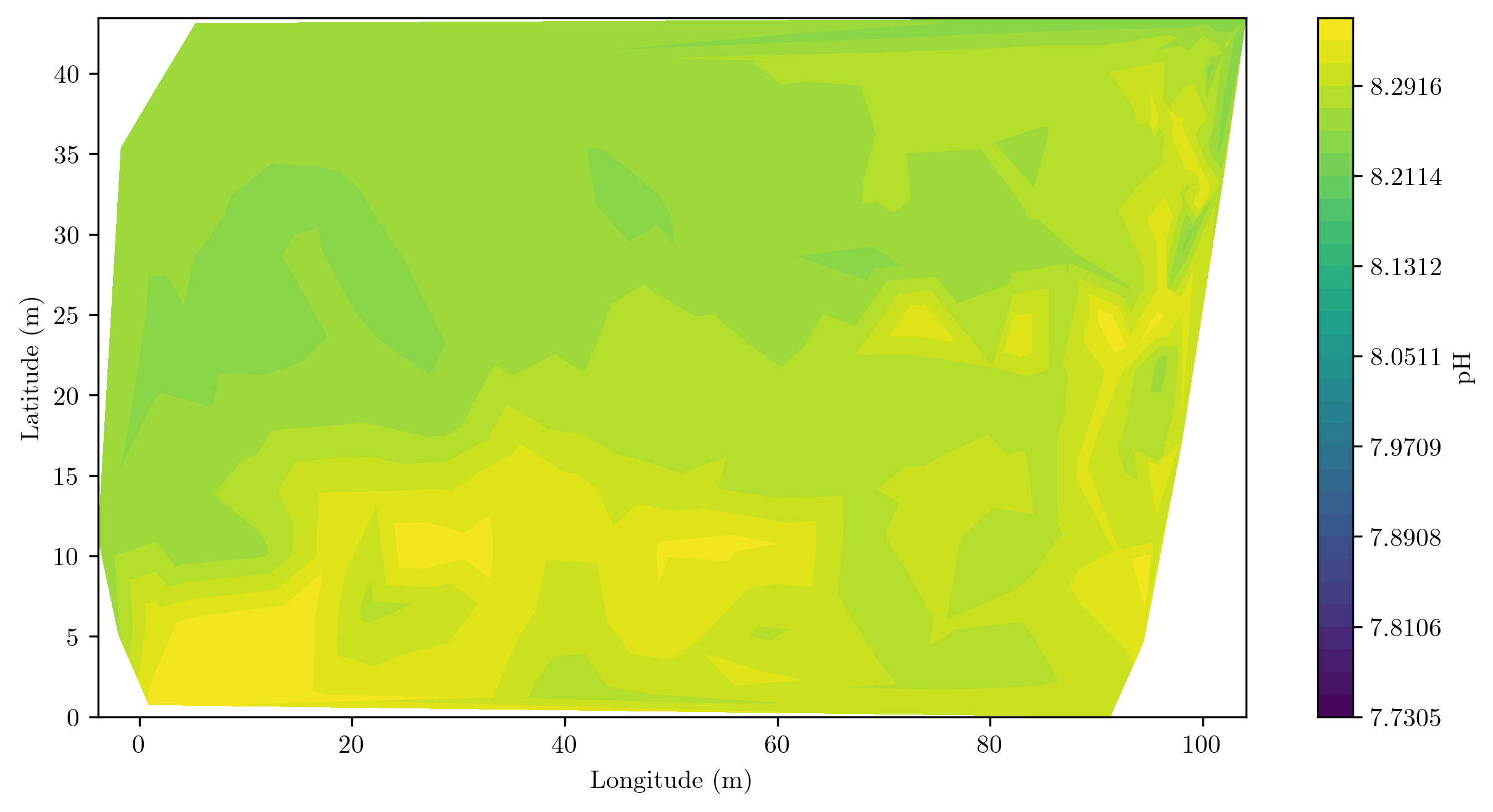}\label{pH5}}		\hspace{0.1cm}
\caption{Interpolated pH measurements.}
\label{pH_all}
\end{figure}

\begin{figure}[H]
    \centering
\subfloat[High-tide at 11:47AM on 08-03-2022.]{\includegraphics[width = 0.8\linewidth]{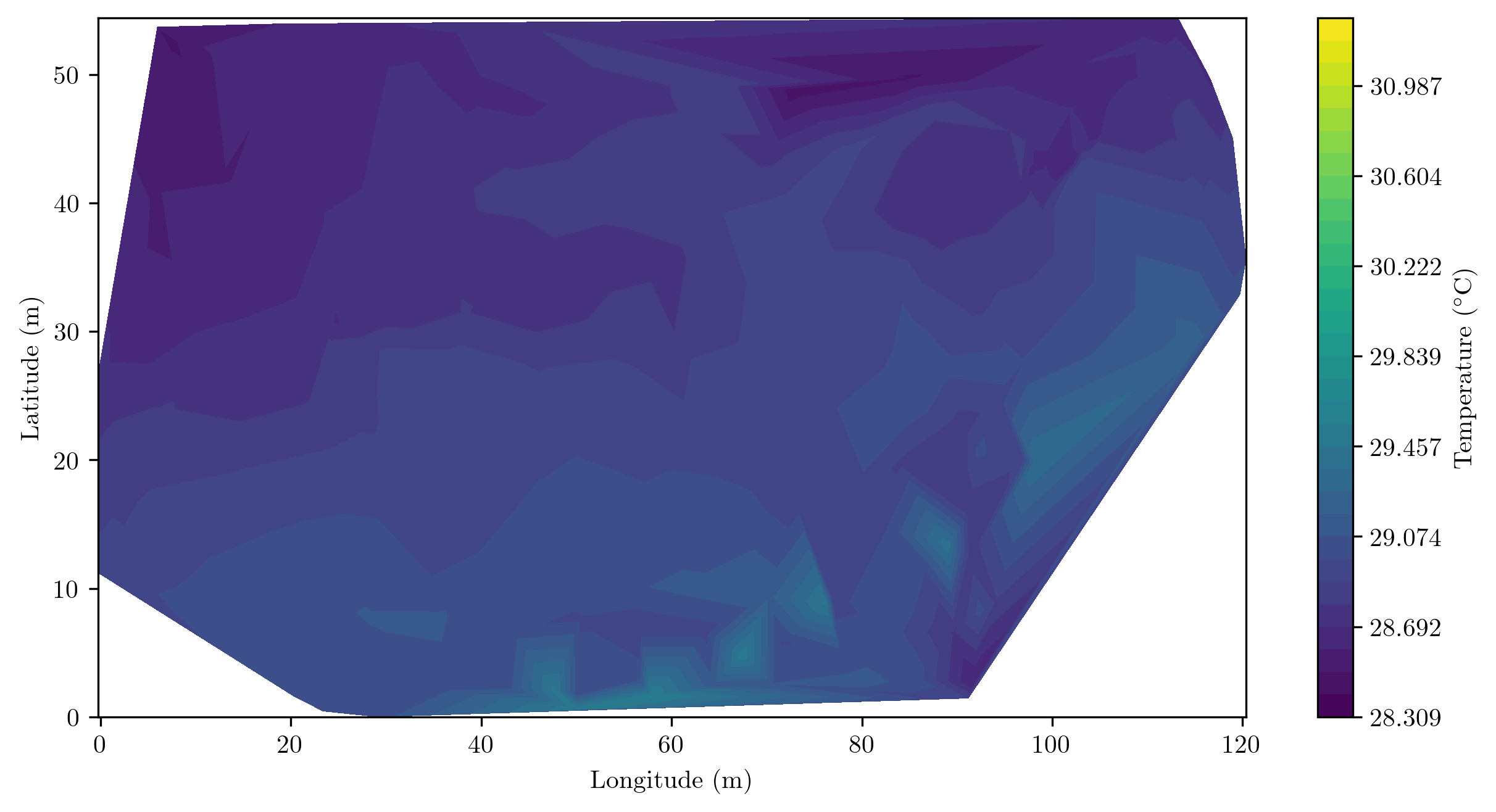}\label{Temperature1}}		\hspace{0.1cm}
\subfloat[High-tide at 1:34PM on 08-03-2022.]{\includegraphics[width = 0.8\linewidth]{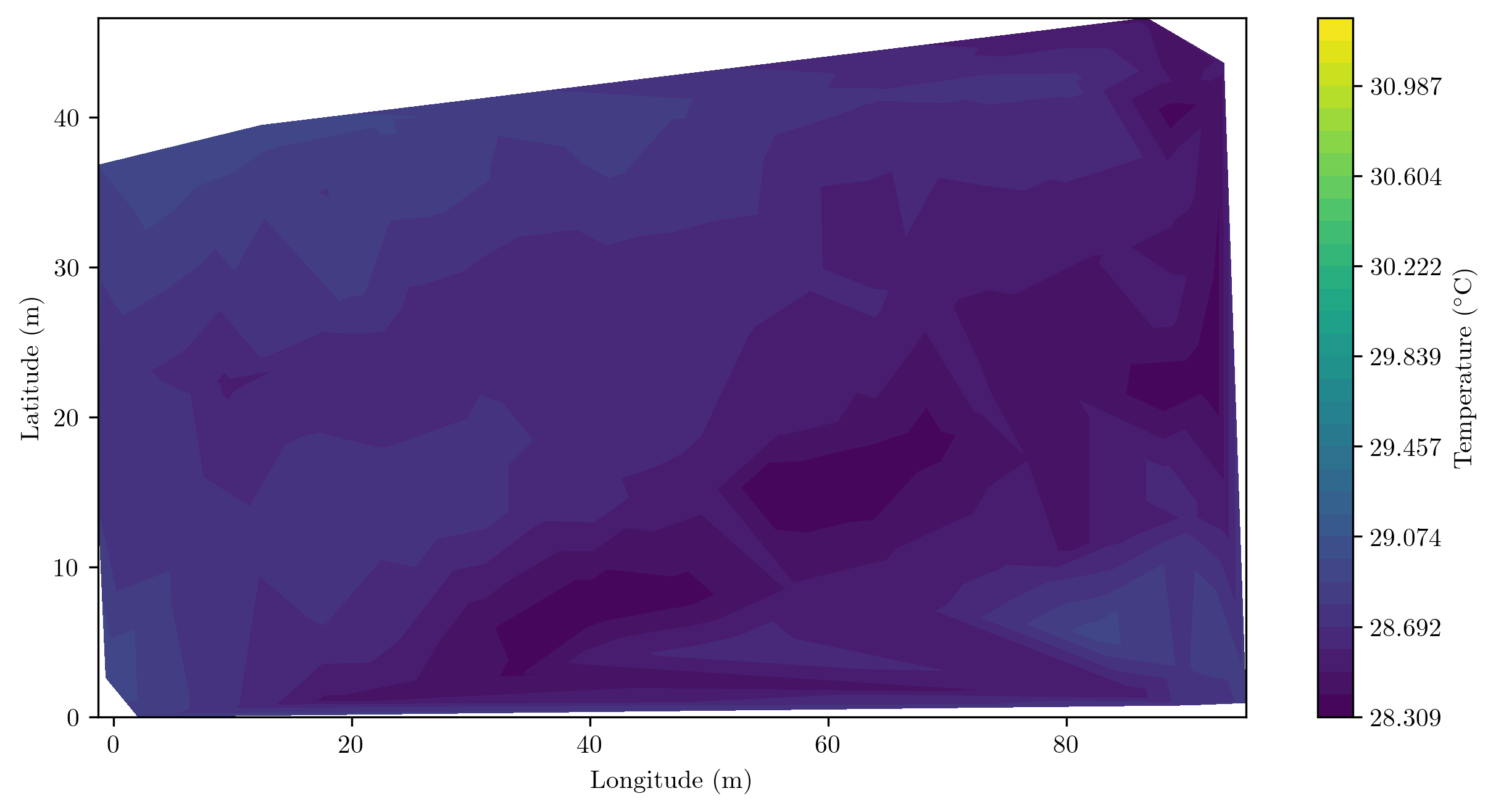}\label{Temperature2}}		\hspace{0.1cm}
\subfloat[Low-tide at 11:07PM on 08-09-2022.]{\includegraphics[width =0.8 \linewidth]{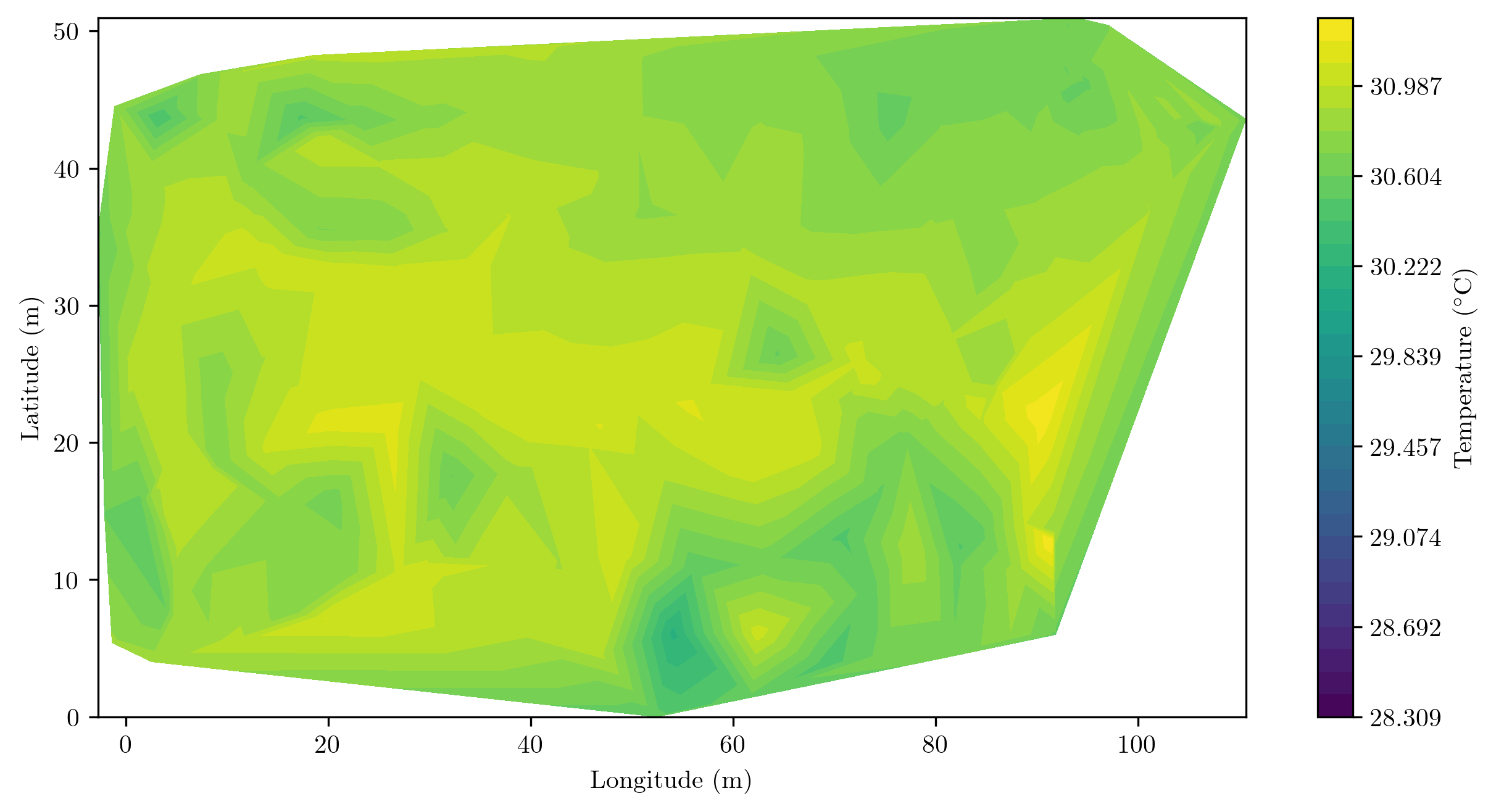}\label{Temperature3}}		\hspace{0.1cm}
\subfloat[Low-tide at 12:03PM on 08-09-2022.]{\includegraphics[width = 0.8 \linewidth]{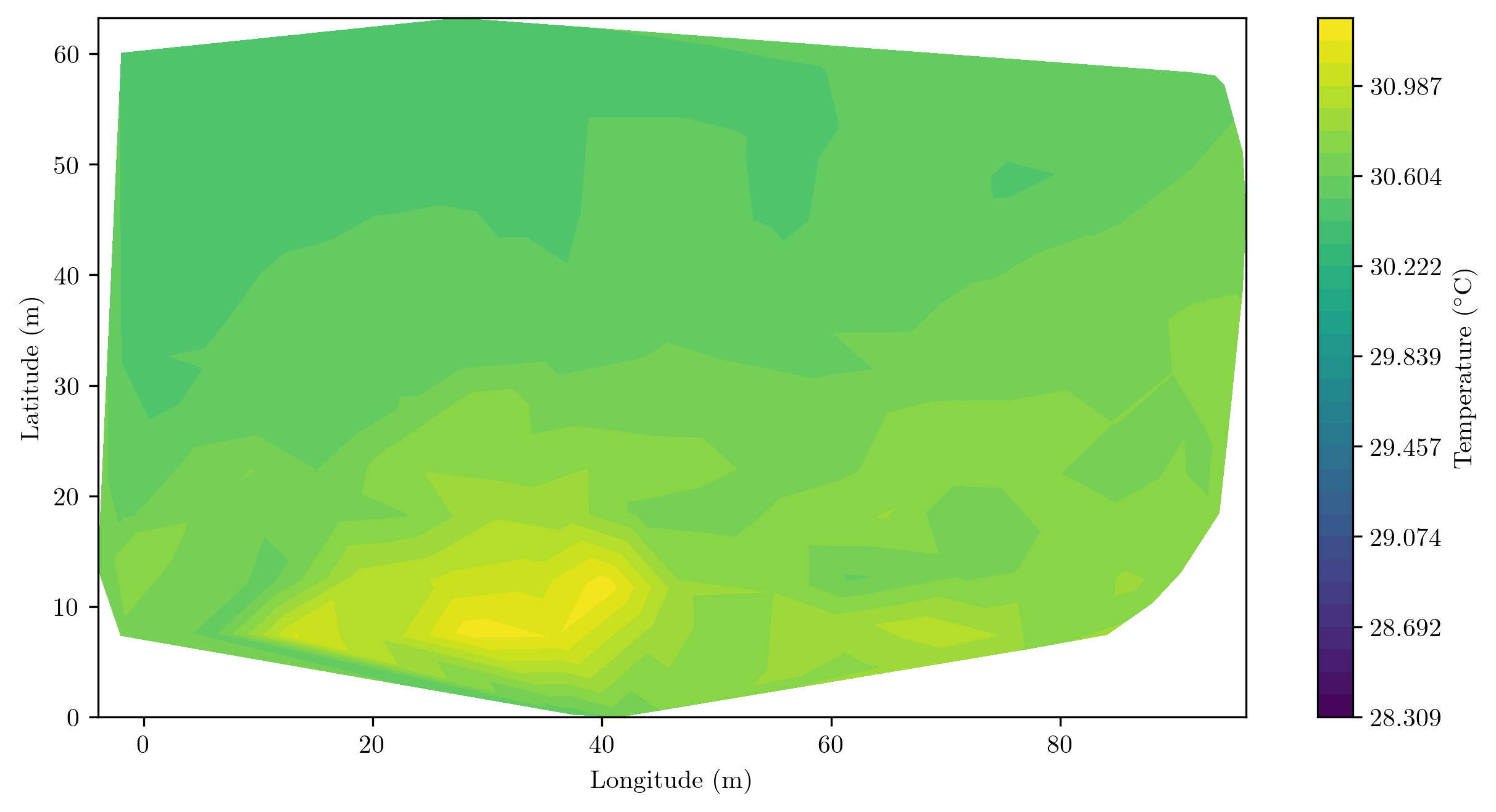}\label{Temperature4}}		\hspace{0.1cm}
\subfloat[Low-tide at 12:19PM on 08-09-2022.]{\includegraphics[width = 0.8 \linewidth]{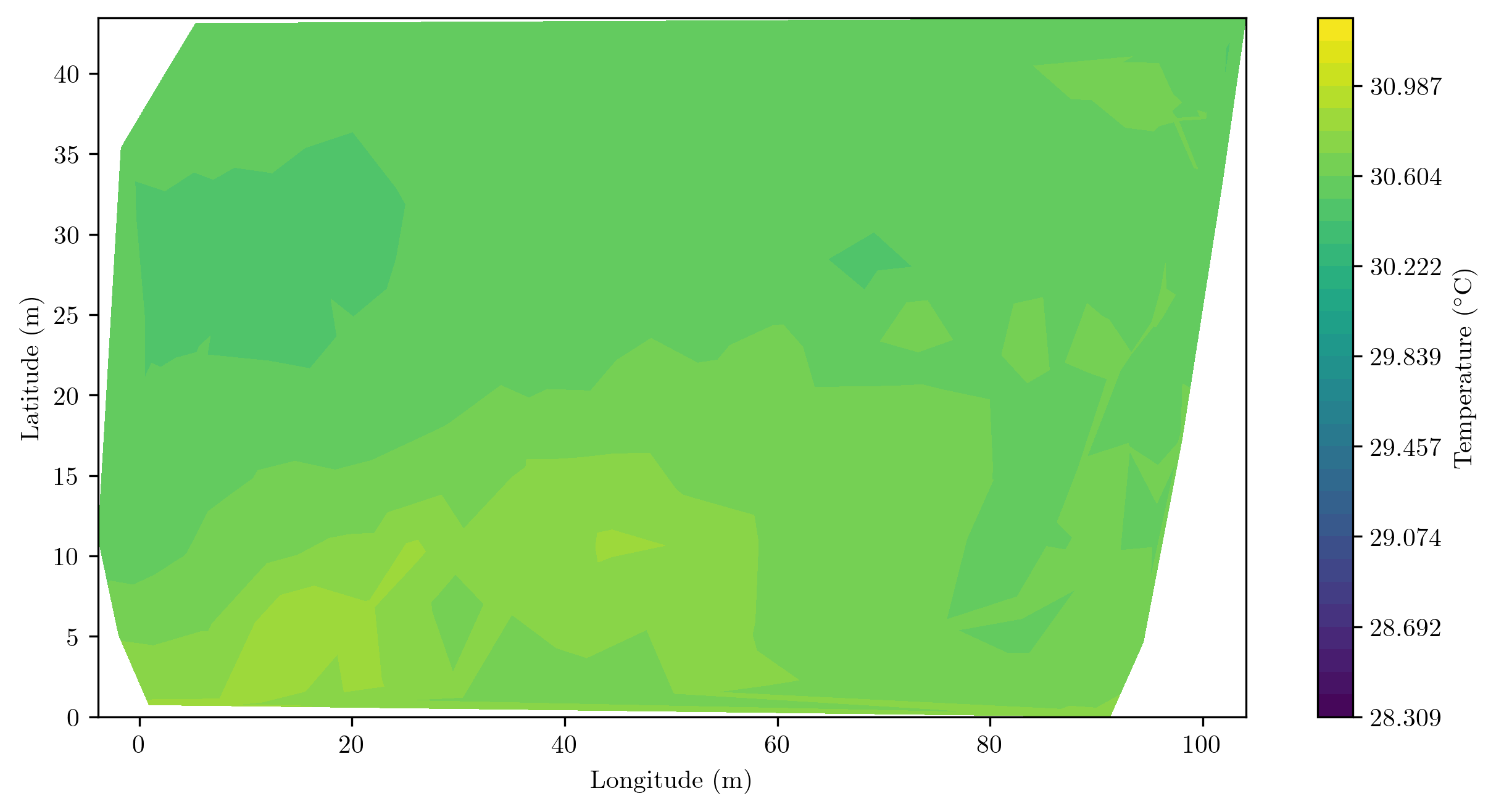}\label{Temperature5}}		\hspace{0.1cm}
\caption{Interpolated temperature measurements.}
\label{temperature_all}
\end{figure}

\begin{figure}[H]
    \centering
\subfloat[High-tide at 11:47AM on 08-03-2022.]{\includegraphics[width = 0.8\linewidth]{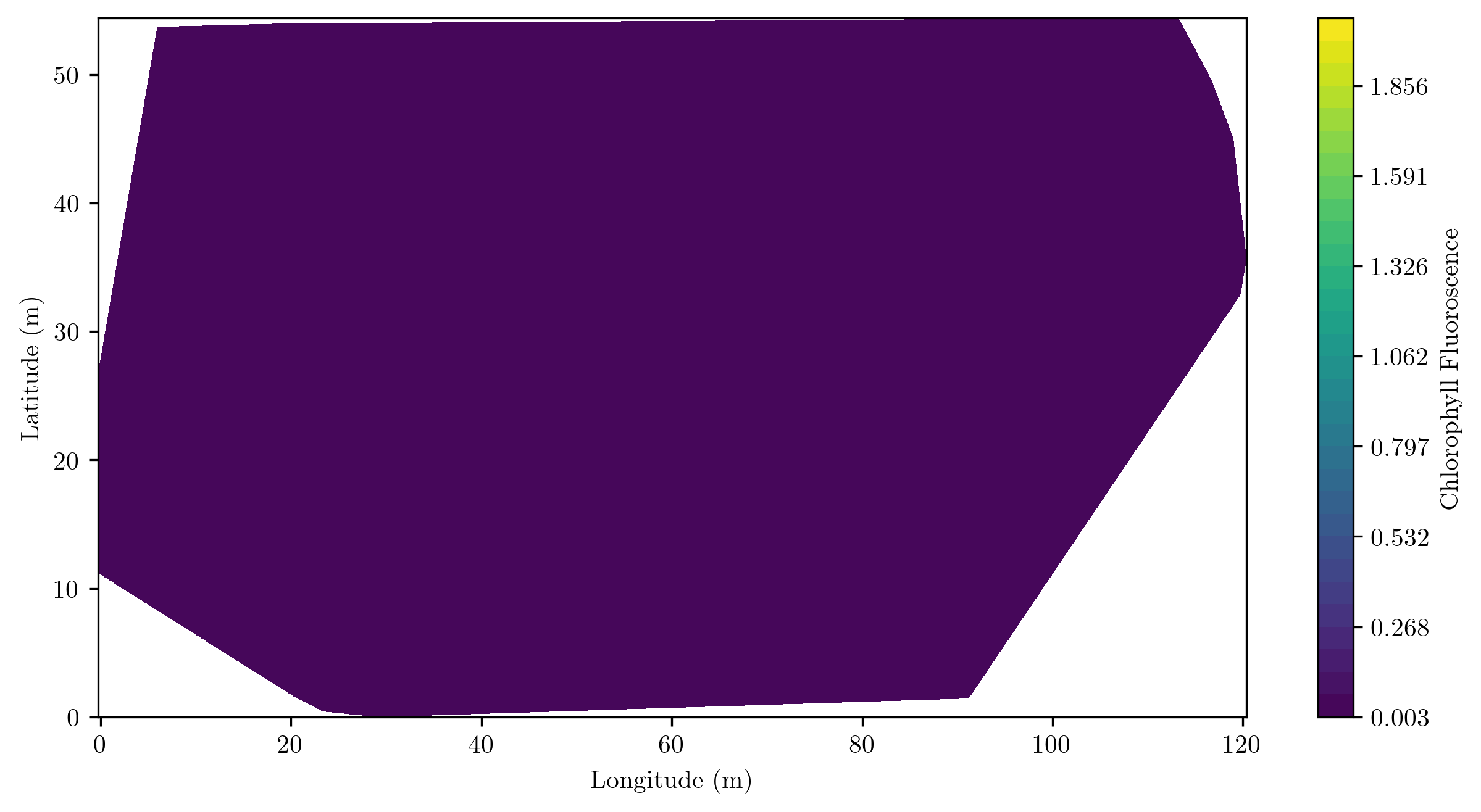}\label{Chlorophyll1}}		\hspace{0.1cm}
\subfloat[High-tide at 1:34PM on 08-03-2022.]{\includegraphics[width = 0.8 \linewidth]{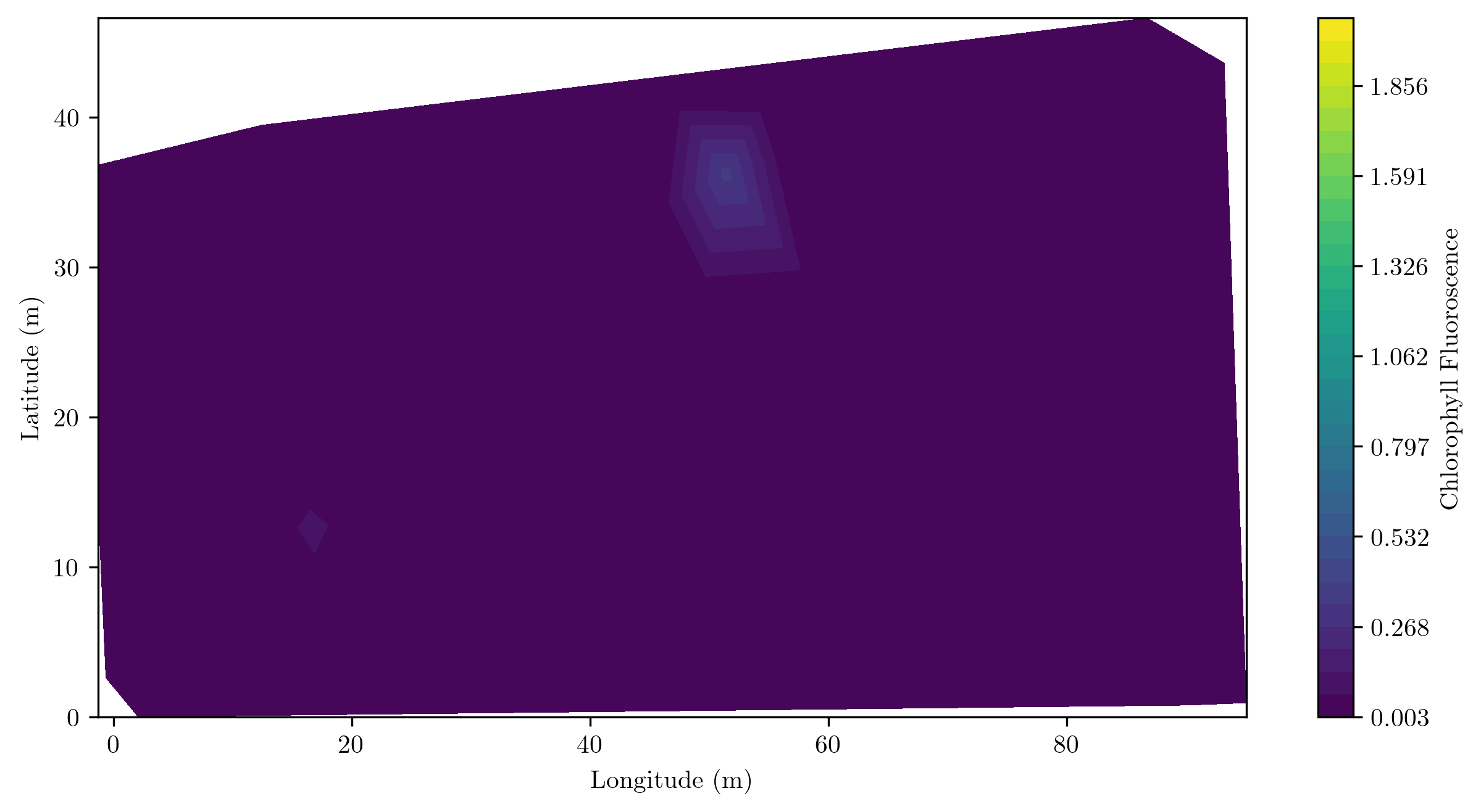}\label{Chlorophyll2}}		\hspace{0.1cm}
\subfloat[Low-tide at 11:07PM on 08-09-2022.]{\includegraphics[width =0.8 \linewidth]{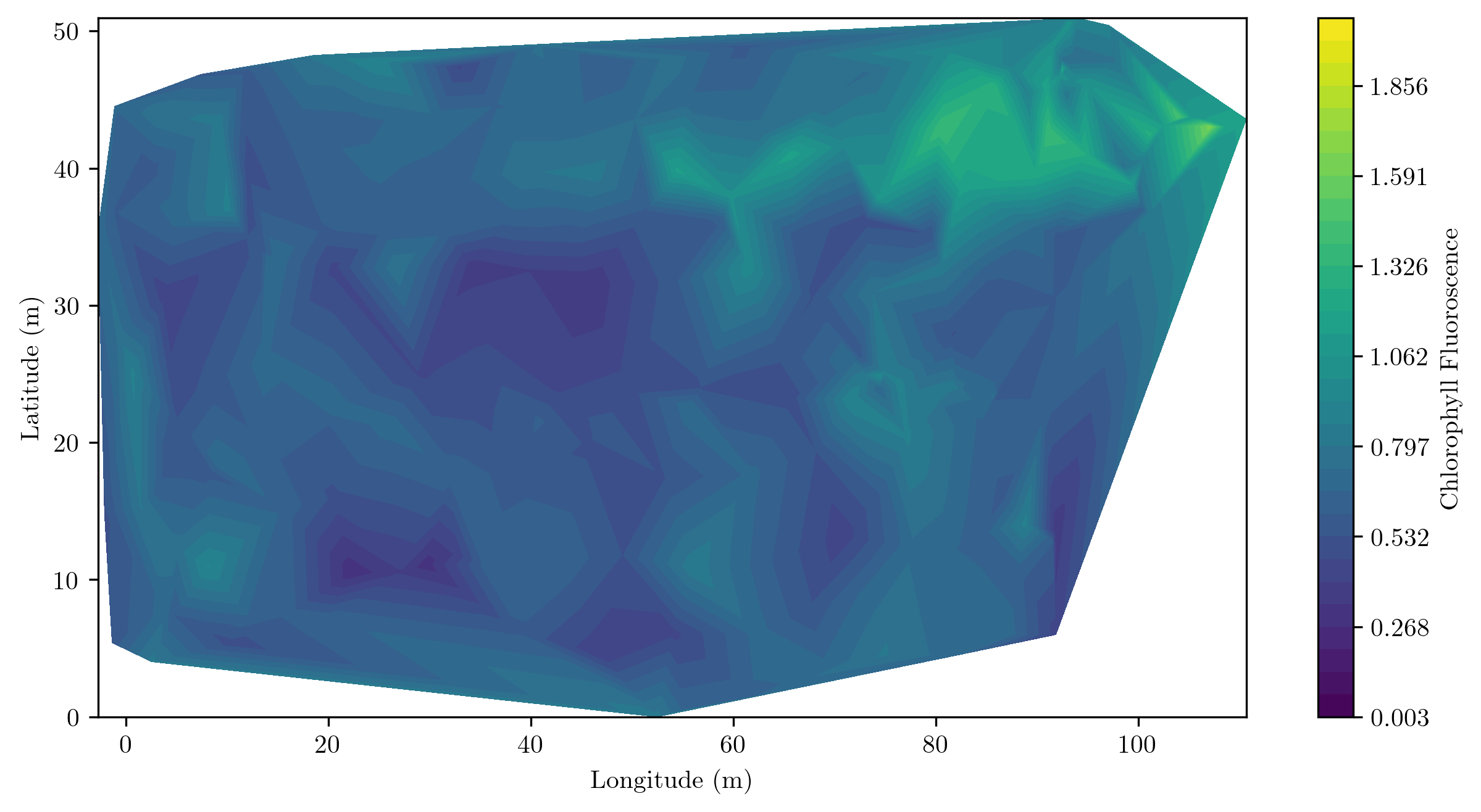}\label{Chlorophyll3}}		\hspace{0.1cm}
\subfloat[Low-tide at 12:03PM on 08-09-2022.]{\includegraphics[width = 0.8 \linewidth]{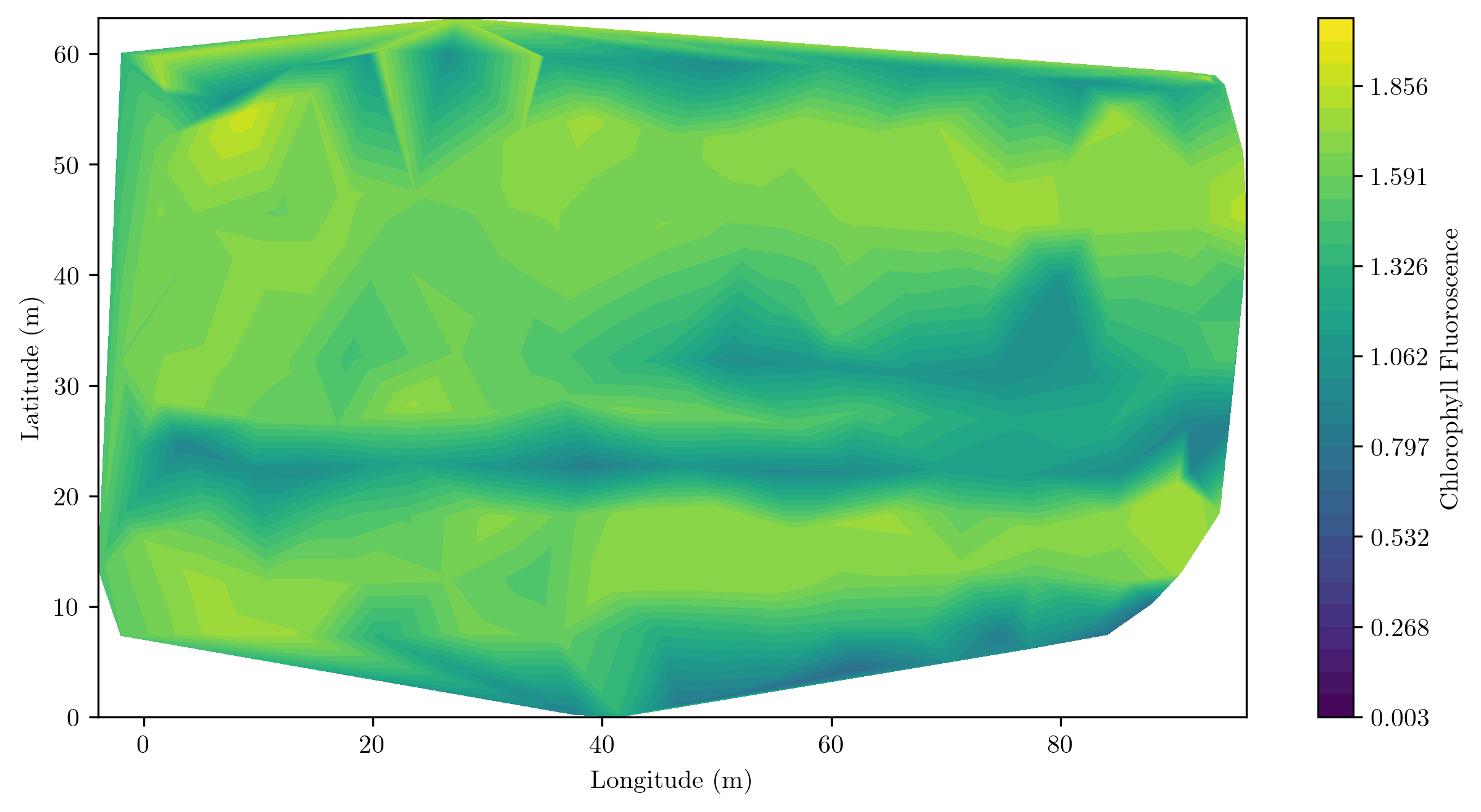}\label{Chlorophyll4}}		\hspace{0.1cm}
\subfloat[Low-tide at 12:19PM on 08-09-2022.]{\includegraphics[width = 0.8 \linewidth]{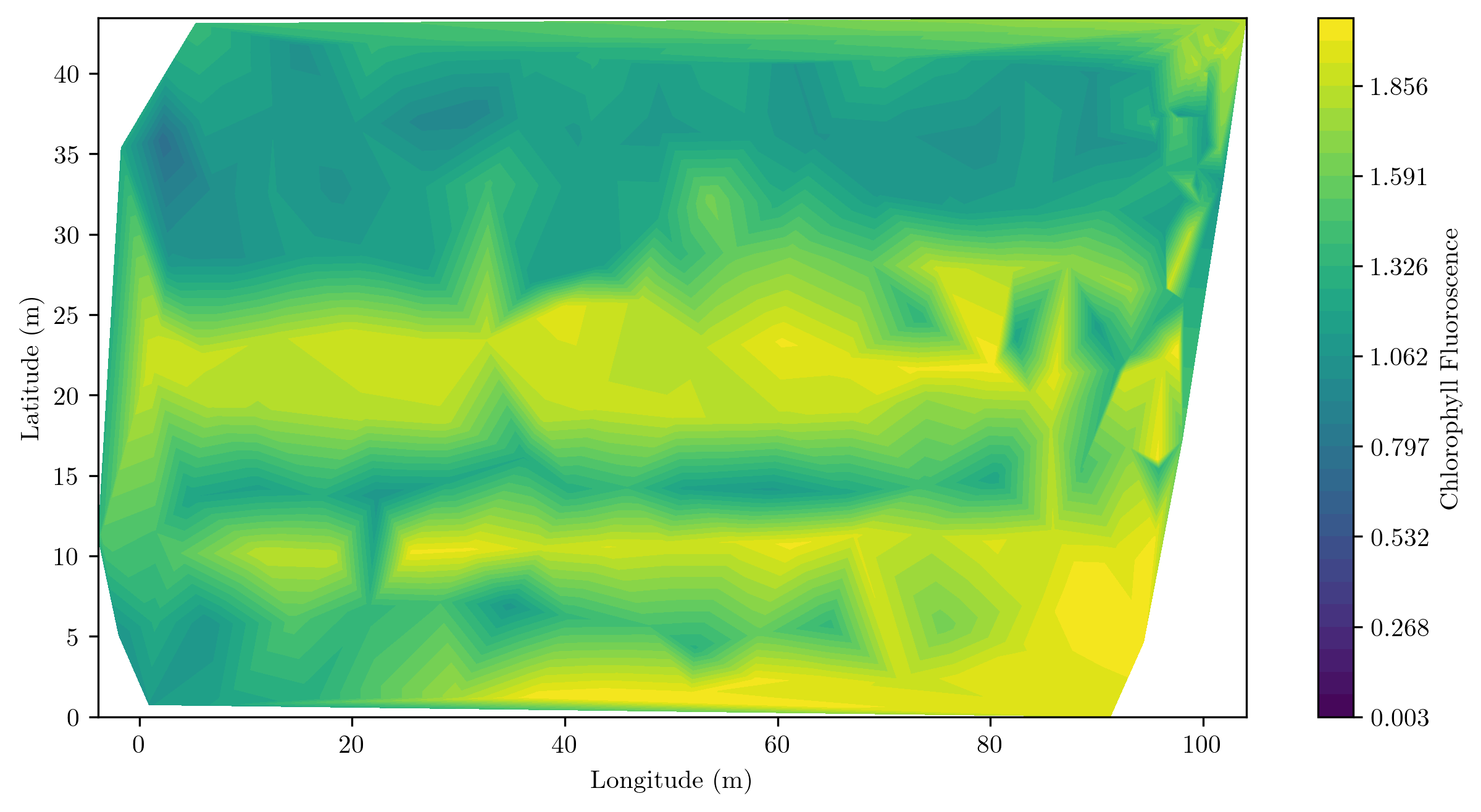}\label{Chlorophyll5}}		\hspace{0.1cm}
\caption{Interpolated chlorophyll-a fluorescence measurements.}
\label{chloro_all}
\end{figure}

Across all datasets, as shown in Figure \ref{pH_all}, pH ranged from 7.73 - 8.35. pH conditions at low-tide were higher than those at high-tide; however, for any given run, the maximum deviation in pH value was 0.18, which is not significant. Next, temperatures in Figure \ref{temperature_all} ranged from \SI{28.3}{^\circ C} to \SI{31.3}{^\circ C}. For both tidal conditions, river water temperatures remained lower than ambient temperatures, staying between \SI{30.2}{^\circ C} - \SI{31.2}{^\circ C} during low-tide when ambient temperatures were around \SI{33}{^\circ C} - \SI{34}{^\circ C}, and between \SI{28.3}{^\circ C} - \SI{29.5}{^\circ C} during high-tide when ambient temperatures were at \SI{31}{^\circ C} - \SI{32}{^\circ C}.  Nitrate concentration levels shown in Figure \ref{nitrate_all}, were between \SI{116.6}{mg/L} - \SI{124.9}{mg/L}, with a maximum deviation of \SI{4.77}{mg/L} for any given run. In general, at high-tide, there were higher nitrate levels.  Pressure remained close to ambient, at {14.7}{psi}, while barometric pressures were around \SI{7.74}{mm Hg} - \SI{8.36}{mm Hg}. ORP levels ranged from \SI{146}{mV} - \SI{262}{mV}. The most deviation in ORP for any run was \SI{40.4}{mV}. Chlorophyll-a Fluorescence in Figure \ref{chloro_all} ranged from \SI{0.003}{RFU} and \SI{2.054}{RFU}, with a maximum change for all runs of \SI{1.34}{RFU}. It should be noted that chlorophyll-a fluorescence measurements were higher on the day of experimentation where water levels were observed to clearer and less murky. The differences in murkiness levels were visible to the eye, however, Secchi disks will be used in future work for more precise quantification of river water visibility. Chlorophyll-a fluorescence measurements had the most variability for this spatial region, while the remaining parameters (pH, temperature, nitrate, pressure, barometric pressure, ORP, sediment concentration) did not show significant variability across the region. This suggests that if spatial relationships between these parameters are to be studied in more detail in the future, ASVs should be deployed to cover a larger region.

\begin{figure}[H]
    \centering
    \includegraphics[width=1\linewidth]{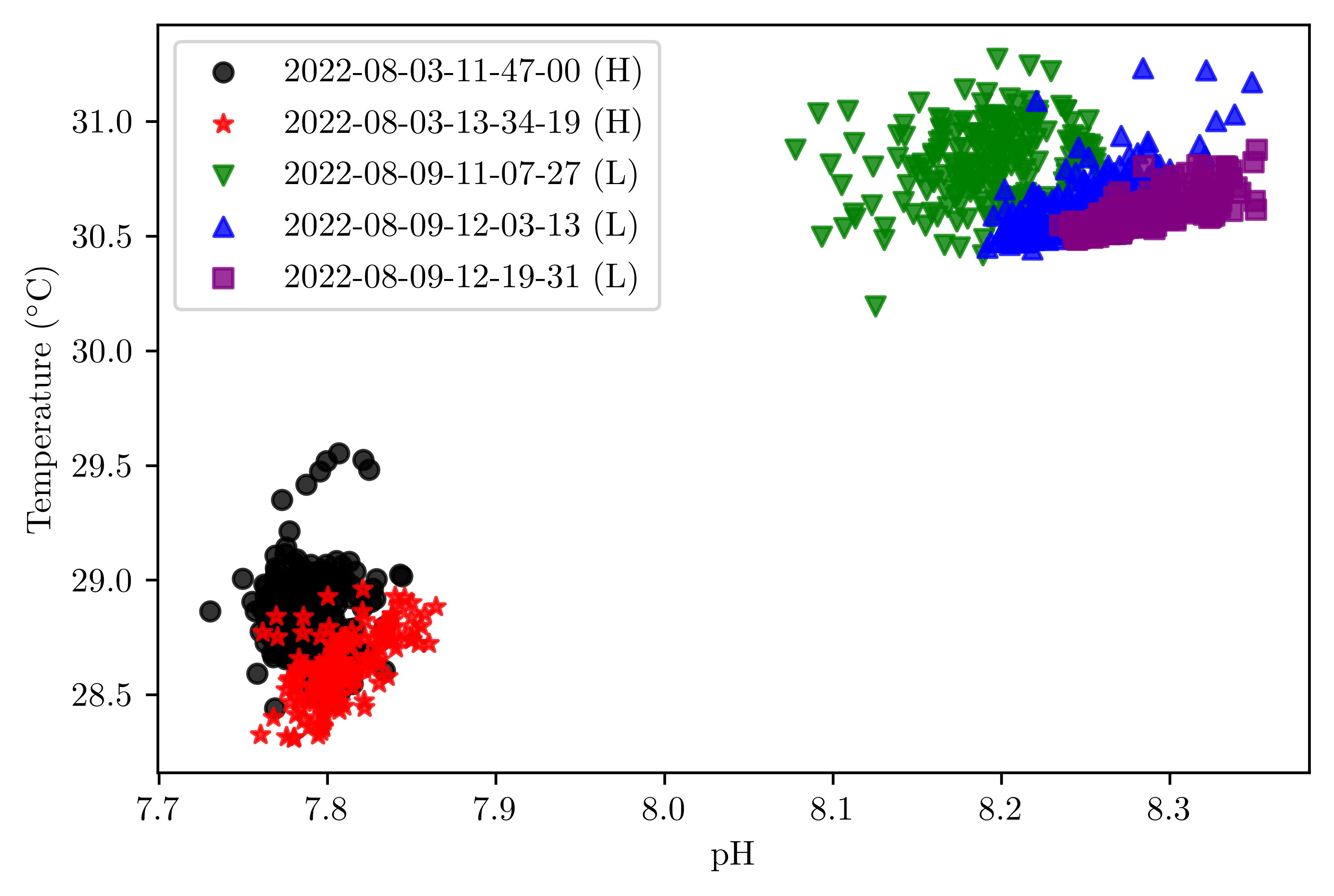}
    \caption{Correlations between pH and temperature, with Pearson correlation coefficient, $R = 0.957$. H represents high-tide, L represents low-tide.}
    \label{pH_vs_temp}
\end{figure}

\begin{figure}[H]
    \centering
    \includegraphics[width=1\linewidth]{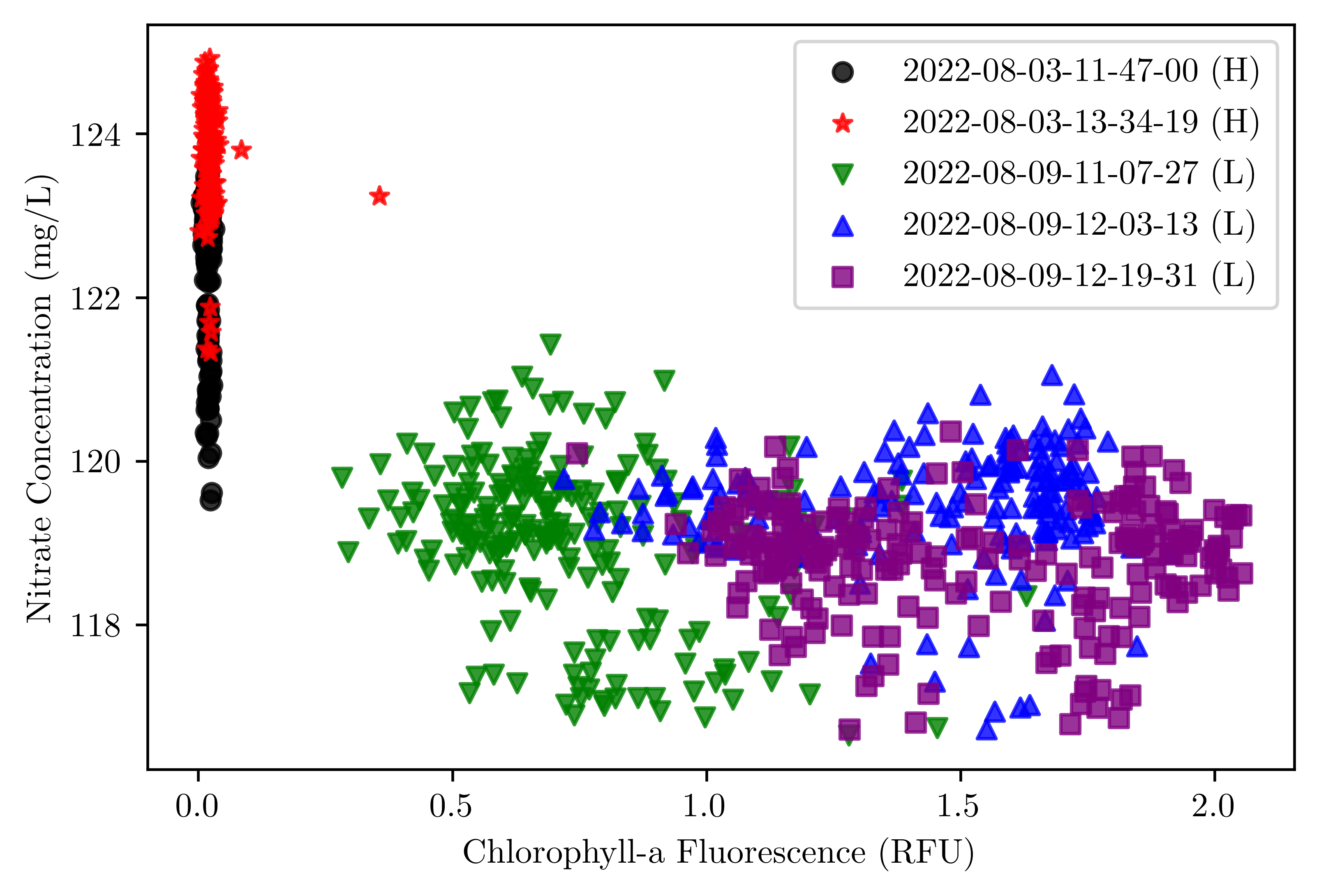}
    \caption{Correlations between chlorophyll-a fluorescence and nitrate concentration, with Pearson correlation coefficient, $R = - 0.77$.}
    \label{chloro_vs_nitrate}
\end{figure} 

\begin{figure}[H]
    \centering
    \includegraphics[width=1\linewidth]{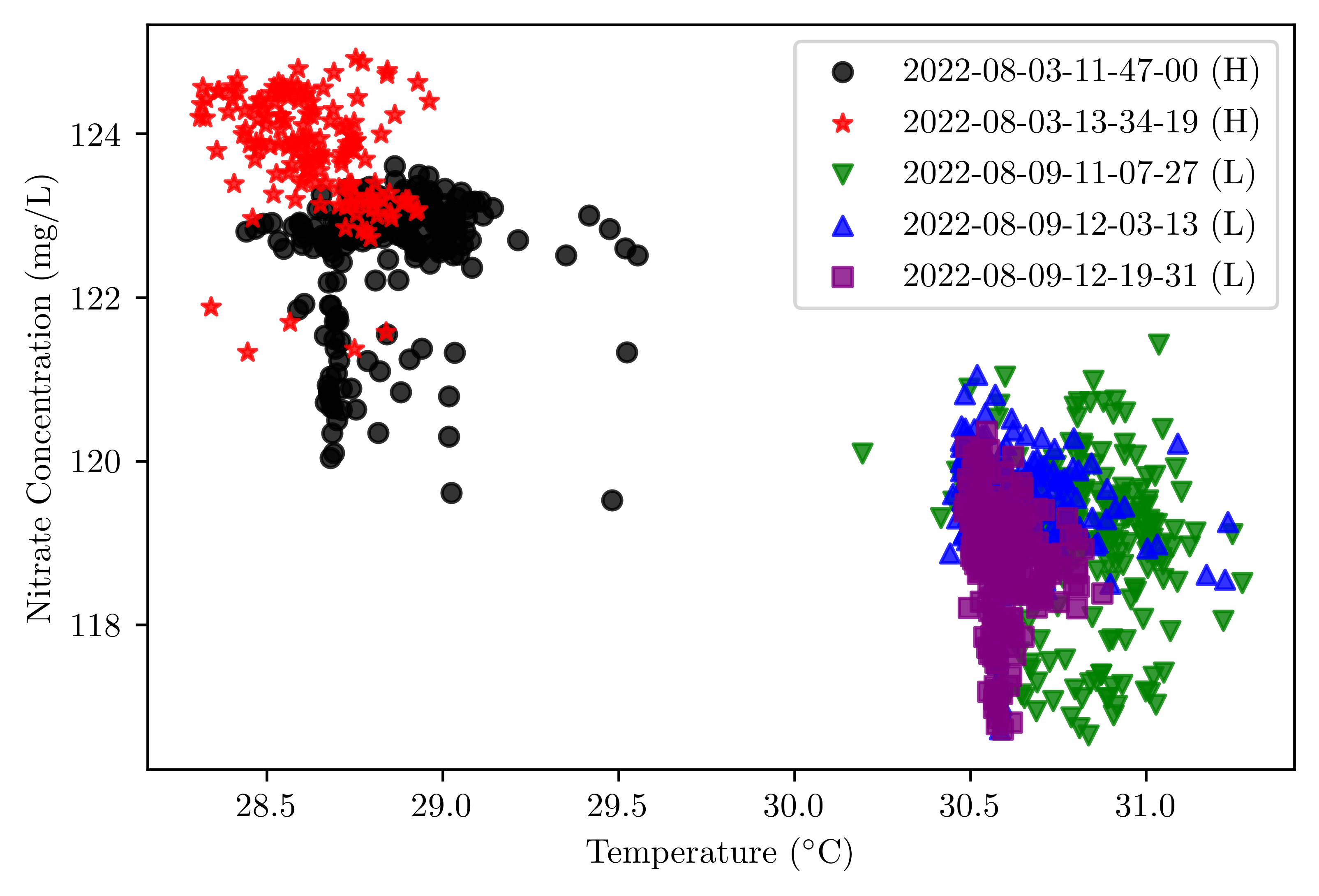}
    \caption{Correlations between temperature and nitrate concentration, with Pearson correlation coefficient, $R =  - 0.903$.}
    \label{temp_vs_nitrate}
\end{figure}

Next, correlations between various pairs of data were also determined. The Pearson correlation coefficient, denoted by R, was used for this. It measures the linear relationship between two sets of data. $ R $ ranges between -1 and +1, where $ R = 1 $ or $ R = -1 $ describes an exact linear relationship. If R is positive, then an increase in the first parameter gives rise to an increase in the second, while if R is negative, then an increase in the first parameter gives rise to a decrease in the second. Figures \ref{pH_vs_temp}, \ref{chloro_vs_nitrate} and \ref{temp_vs_nitrate} illustrate such relationships. 

Figure \ref{pH_vs_temp} shows the relationship between temperature and pH. From this data, $ R = 0.957 $, pH increases with an increase in temperature. This relationship can be explained as follows: with an increase in temperature comes a decrease in dissolution of carbon dioxide into the water. More carbon dioxide introduced in the water means greater river acidity. This means that with increased river water temperature, we have greater pH -- a more basic measurement. For chlorophyll-a fluorescence vs nitrate concentration shown in Figure \ref{chloro_vs_nitrate}, $ R = -0.77 $. Again, the relationship obtained is to as expected. Chlorophyll measurement is a proxy for the existence of phytoplankton. Phytoplankton consumes nitrates, so when algae blooms develop, the level of nitrates in the river decreases. This gives rise to the negative correlation between the two parameters. Finally, nitrate levels as a function of temperature are plotted in Figure \ref{temp_vs_nitrate}. For this, $R  = -0.903$. This correlation agrees with the relationship of that in rivers, where nitrate levels tend to be lower during the winter and higher during summer months. 

\section{Conclusions and Future Work}
Autonomous Surface Vehicles were deployed in the Schuylkill river in Philadelphia. A custom made ASV surveyed the river autonomously in a lawn-mower pattern and measured pH, temperature, nitrate, pressure, barometric pressure, ORP, and chlorophyll-a fluorescence, sediment concentration, and bathymetry data over a \SI{90}{m} $\times$ \SI{40}{m} region. A Clearpath Heron was used to acquire additional depth measurements across a \SI{100}{m} $\times$ \SI{75}{m} region. Data were acquired under high-tide and low-tide conditions, on two different days, to allow for measurements during similar times of the day, and therefore similar ambient temperatures. Experimental results show that there are clear trends among the same tidal conditions, namely, high-tide data and low-tide data. Collectively, when analyzing all high-tide and low-tide data together, Pearson correlations between pH, temperature, nitrate, and chlorophyll-a fluorescence meet expectations, while correlations between remaining parameters are either not conclusive with current data or do not exist. 2D reconstructions of the riverbed indicate that there is some asymmetry in the river bed, due to the meandering of the river before and after the region of measurement.

Future surveys will target river responses to floods, to understand regime shifts in hydrodynamics and water chemistry and to document erosion hotspots that threaten infrastructure. Long-term surveys will examine annual cycles of sediment, salt, and flow in the river, and changes in the bathymetry that may influence flood capacity of the Lower Schuylkill River. Experiments will also be repeated under additional tidal conditions \emph{i.e.} during hours between peak high-tide and low-tide, and over larger spatial regions. In addition, measurements will be acquired at various depths, to generate 3D reconstructions of the parameters acquired in this study. This data would be used to better understand the fluid flow and sediment dynamics in the river, as well as develop adaptive sampling strategies.

\bibliographystyle{IEEEtran}
\bibliography{refs}
\vspace{12pt}

\end{document}